\documentclass[twoside]{article}

\usepackage[accepted]{aistats2023}
\usepackage[round]{natbib}

\usepackage{mathtools} 
\usepackage{booktabs}  
\usepackage{hyperref}  
\hypersetup{
    colorlinks = true,
    citecolor  = blue,
    linkcolor  = blue
    }
\usepackage{url}                      
\usepackage{amsmath,amssymb,amsfonts} 
\usepackage{graphicx}                 
\usepackage{enumitem}
\usepackage{algpseudocode}

\usepackage{algorithm}
\usepackage{array}
\usepackage{float}

\newcommand\ie{i.\,e.}
\newcommand\eg{e.\,g.}

\begin{document}

%
\runningtitle{Estimating Conditional Average Treatment Effects with Missing Treatment Information}

%

\twocolumn[

\aistatstitle{Estimating Conditional Average Treatment Effects \\ with Missing Treatment Information}

\aistatsauthor{ Milan Kuzmanovic \And Tobias Hatt \And  Stefan Feuerriegel }

\aistatsaddress{ ETH Zurich \And  ETH Zurich \And ETH Zurich, LMU Munich } ]

\begin{abstract}
Estimating conditional average treatment effects (CATE) is challenging, especially when treatment information is missing. Although this is a widespread problem in practice, CATE estimation with missing treatments has received little attention. In this paper, we analyze CATE estimation in the setting with missing treatments where unique challenges arise in the form of covariate shifts. We identify two covariate shifts in our setting: (i)~a covariate shift between the treated and control population; and (ii)~a covariate shift between the observed and missing treatment population. We first theoretically show the effect of these covariate shifts by deriving a generalization bound for estimating CATE in our setting with missing treatments. Then, motivated by our bound, we develop the missing treatment representation network~(MTRNet), a novel CATE estimation algorithm that learns a balanced representation of covariates using domain adaptation. By using balanced representations, \mbox{MTRNet} provides more reliable CATE estimates in the covariate domains where the data are not fully observed. In various experiments with semi-synthetic and real-world data, we show that our algorithm improves over the state-of-the-art by a substantial margin.  
\end{abstract}

\section{INTRODUCTION}
\label{sec:intro}


Estimating conditional average treatment effects~(CATE) is crucial for decision-making in many application domains such as economics \citep{Smith2005, Baum2015}, marketing \citep{Wang2015, Li2016, hatt2020early}, and medicine \citep{Alaa2017}. For example, a doctor deciding on a personalized treatment plan based on patient characteristics. Extensive work focuses on using machine learning to estimate CATE \citep[\eg, ][]{Shalit2017, Alaa2018, Yoon2018, Athey2019, frauen2022estimating}. However, existing work has given little attention to settings where treatment information is missing. 


Missing treatment information is common in many real-world applications \citep{Kennedy2020}. For instance, \citet{Zhang2016} describe the Consortium on Safe Labor study, where the question of interest is the causal effect of mothers' body mass index~(BMI) on infants' weight. In this study, BMI was missing for about half of the subjects. Another example is provided by \citet{Ahn2011}, where the authors analyze the effect of physical activity on colorectal cancer using data from the Molecular Epidemiology of Colorectal Cancer study. However, information on physical activity was missing for around 20\,\% of the subjects. Further, \citet{Molinari2010} gives numerous examples of missing treatment information in survey settings. Missing treatment information can create additional challenges for treatment effect estimation. Motivated by needs in practice, the question is how one can reliably estimate CATE even when treatment information is missing.


In this paper, we analyze the problem of estimating CATE with missing treatment information. We consider a causal structure where both treatment and treatment missingness are affected by covariates. In such setting, we have two covariate shifts: (i)~a covariate shift between the treated and control population; and (ii)~a covariate shift between the observed and missing treatment population. These covariate shifts increase CATE estimation error in covariate domains where we lack fully observed data. For instance, if low-income patients are reluctant to share information about their treatment, they can be largely underrepresented in the observed treatment population, and, hence, CATE estimation for low-income patients might be unreliable due to the lack of observed treatment data. We theoretically show the effect of these covariate shifts by deriving a generalization bound for estimating CATE in our setting with missing treatments. Our derivation shows that the expected CATE estimation error is bounded by the sum of (i)~the standard estimation error; (ii)~the distance between the covariate distributions of the treated and control population; and (iii)~the distance between the covariate distributions of the observed and missing treatment population. 


Our generalization bound reveals that we need to account for the two covariate shifts when estimating CATE in our setting with missing treatments. Motivated by our bound, we propose the \emph{missing treatment representation network}~(MTRNet), a novel CATE estimation algorithm for our setting with missing treatments. MTRNet makes use of representation learning \citep{Bengio2013} and domain adaptation \citep{Ganin2016} to address the covariate shifts while aiming at a low CATE estimation error. In particular, MTRNet uses adversarial learning to learn a balanced representation of covariates which is neither predictive of treatment nor of treatment missingness. By using balanced representations, we reduce the CATE estimation error in domains that have different covariate distributions than the one in which we fully observe data, and, thus, we improve the overall performance. In various experiments with semi-synthetic and real-world data, we demonstrate that our MTRNet yields superior CATE estimates in our setting with missing treatment information compared to the state-of-the-art. 


We list our main \textbf{contributions}\footnote{Code available at: \\ \url{https://github.com/mkuzma96/MTRNet}} as follows:
\begin{enumerate} 
\item We analyze the problem of estimating CATE with missing treatment information. To the best of our knowledge, existing literature on CATE estimation has previously overlooked this setting.  
\item We derive a generalization bound that shows different sources of error that we need to account for when estimating CATE in the setting with missing treatments. 
\item We develop MTRNet, a novel CATE estimation algorithm based on our generalization bound. Across various experiments, we demonstrate that MTRNet provides superior CATE estimates in our setting with missing treatments compared to the state-of-the-art. 
\end{enumerate}

\section{RELATED WORK}
\label{sec:relwork}

We review two streams in the literature that are particularly relevant to our problem (\ie, CATE estimation with missing treatments): (i)~methods for average treatment effect~(ATE) estimation with missing treatments, and (ii)~methods for CATE estimation in the standard setting that address the covariate shift between the treated and control population. 


\textbf{(i)~ATE estimation with missing treatments.} Only a few methods have been developed for estimating treatment effects in the setting with missing treatment information. These methods primarily focus on identification and estimation of average treatment effects. \citet{Williamson2012} proposed a doubly robust augmented inverse probability weighted estimator for ATE that deals with both confounding and missing treatments. \citet{Zhang2016} combined standard causal inference and missing data models to create a triply robust estimator for ATE. Both estimators are semi-parametric and thus offer certain robustness to misspecification; however, they are restricted to standard parametric models as nuisance functions. \citet{Kennedy2020} proposed a nonparametric estimator for ATE in the missing treatment setting that can incorporate flexible machine learning models for nuisance functions. 


The major difference between the existing literature on treatment effect estimation with missing treatments \citep{Williamson2012,Zhang2016,Kennedy2020} and our work is that our focus is not on ATE but on CATE estimation. In fact, the existing methods focus only on identification and direct estimation of ATE. As such, they cannot be straightforwardly adapted to CATE estimation and are thus \emph{not} applicable to our setting. To the best of our knowledge, we are the first to study CATE estimation with missing treatment information.   


\textbf{(ii)~CATE estimation in the standard setting.} Numerous methods have been proposed for estimating CATE \citep[\eg, ][]{Alaa2018, Yoon2018, Athey2019}. Here, we focus on methods that address the covariate shift between the treated and control population, as our work deals with covariate shifts for CATE estimation as well. \citet{Johansson2016} were the first to identify the covariate shift problem when estimating CATE. In order to account for the covariate shift, the authors propose an algorithm that learns a balanced representation of covariates by enforcing domain invariance through distributional distances. \citet{Shalit2017} extended their work by deriving a more flexible family of algorithms for this task. The authors also provide an intuitive generalization bound for CATE estimation that theoretically shows the effect of the covariate shift. Building on top of these works, other methods were proposed for addressing the covariate shift between the treated and control population, some of which include learning weighted representations \citep{Johansson2018, Assaad2021, hatt2022combining} and learning overlapping representations \citep{Zhang2020}.


In our work, we also address the covariate shift between the treated and control population since we have a CATE estimation problem. However, we consider a more general setting with missing treatments where we identify an additional covariate shift between the observed and missing treatment population that needs to be accounted for. This covariate shift, as well as the setting with missing treatments in general, was not studied by any prior work on CATE estimation. Moreover, due to having two covariate shifts, our proposed algorithm is designed to learn a covariate representation that is balanced over multiple domains. This requires a tailored approach that differentiates from the above methods. 

\section{PROBLEM SETUP} 
\label{sec:setup}


Let $\mathcal{T} = \{0,1\}$ denote whether a treatment is \emph{applied}, and let $\mathcal{R} = \{0,1\}$ denote whether the treatment information is \emph{observed} (or missing). Further, we refer to a covariate space $\mathcal{X} \subseteq \mathbb{R}^d$ and an outcome space $\mathcal{Y} \subseteq \mathbb{R}$. We describe the outcomes of different treatments using the Rubin-Neyman potential outcomes framework \citep{Rubin2005}. We assume a distribution $p(t, r, x, y_0, y_1)$ with the following variables: treatment assignment $T \in \mathcal{T}$, treatment missingness $R \in \mathcal{R}$, covariates $X \in \mathcal{X}$, and potential outcomes $Y_0, Y_1 \in \mathcal{Y}$. We observe only one potential outcome, \ie, we observe $Y \in \mathcal{Y}$, where $Y = Y_0$ or $Y = Y_1$, depending on the assigned treatment $T = t$. The observed potential outcome corresponding to the assigned treatment $t$ is called the factual outcome, and the unobserved potential outcome corresponding to the other treatment possibility (\ie, $1-t$) is called the counterfactual outcome. We have data for $n$ individuals given by $\mathcal{D} =  \{(t_i, r_i, x_i, y_i)\}_{i=1}^{n}$, where $t_i$ is observed only if $r_i = 1$. That is, some of treatment information is missing. 

\begin{figure}[tbp]
\centerline{\includegraphics[width= 5cm]{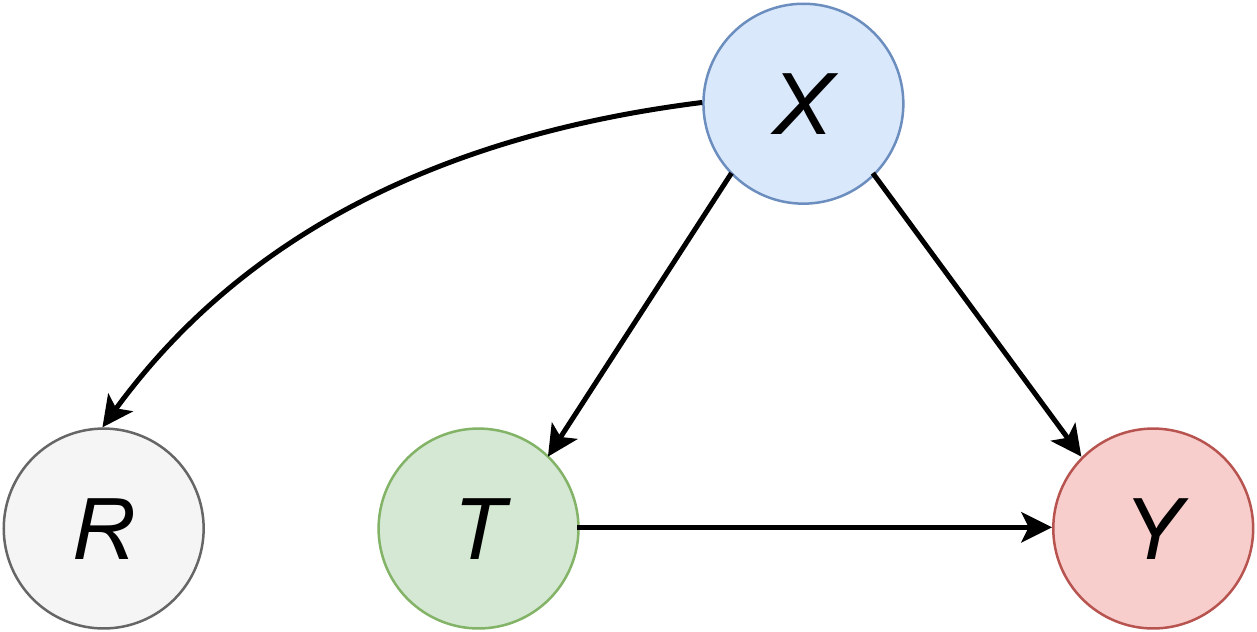}}
\caption{Overview of the causal structure in our setting.}
\label{fig:causal_structure}
\end{figure}

Our objective is to estimate the conditional average treatment effect~(CATE)\footnote{Also known as the individualized treatment effect (ITE).} for an individual with covariates $X = x$ from data $\mathcal{D}$ with missing treatment information. This is given by
\begin{equation}
\tau(x) := \mathbb{E} \, [ \, Y_1 - Y_0 \mid X = x \, ].
\end{equation}
We make the following assumptions about our setting with missing treatments (the causal structure of our problem is illustrated in Fig.~\hyperref[fig:causal_structure]{1}):

\textbf{Assumption 1} \emph{(Consistency, $T$-Positivity, $T$-Ignorability)}.
\begin{enumerate}[label=(\roman*)]
\label{ass:trt}
\item \emph{$Y = Y_0$ if $T = 0$, and $Y = Y_1$ if $T = 1$ (Consistency);}
\item \emph{\mbox{$0 < p(T = 1 \mid X = x) < 1$ if $p(x) > 0 $ ($T$-Positivity);}}
\item \emph{$Y_0, Y_1 \perp \!\!\! \perp T \mid X = x \, \, $ ($T$-Ignorability).}
\end{enumerate}

Assumption~1 are the standard assumptions for identification of treatment effects from data. $T$-Ignorability is often referred to as `no hidden confounders' assumption\footnote{Also known as exchangeability \citep{melnychuk2022normalizing} or strong ignorability \citep{hatt2021generalizing}}, meaning that all variables that affect both treatment $T$ and potential outcomes $Y_0$ and $Y_1$ are measured in covariates $X$. 

\textbf{Assumption 2} \emph{($R$-Positivity, $R$-Ignorability)}.
\begin{enumerate}[label=(\roman*)]
\label{ass:miss}
\item \emph{\mbox{$0 < p(R = 1 \mid X = x) < 1$ if $p(x) > 0 $ ($R$-Positivity);}}
\item \emph{$ R \perp \!\!\! \perp T, Y_0, Y_1 \mid X = x \, \, $ ($R$-Ignorability).}
\end{enumerate}

Assumption~2 corresponds to a standard variant of missing at random~(MAR) assumption: the missingness depends only on the fully observed part of the data (in our case on the covariates $X$). Together, Assumption~1 and Assumption~2 allow for identification of treatment effects from data with missing treatment information \citep{Zhang2016, Kennedy2020}. 

Note that Assumption~1 and Assumption~2 can also hold in cases where covariates $X$ are: (i)~independent of treatment $T$, (ii)~independent of treatment missingness $R$, or (iii)~independent of both treatment $T$ and treatment missingness $R$. Hence, our setting is not restrictive in the sense that we require covariates to affect both treatment and treatment missingness, rather, it is more general and also applicable in cases where covariates do not affect one of the two, or do not affect either of the two.

The fundamental problem of causal inference is that counterfactual outcomes (\ie, outcomes under a different treatment than the one assigned) are unobserved. Additionally, in our setting, we also have missing treatment information. Unobserved counterfactual outcomes and missing treatments preclude direct estimation of CATE from data. However, under Assumption~1, we have $\mathbb{E} \, [ \, Y_t \mid X = x \, ] = \mathbb{E} \, [ \, Y \mid X = x, T = t \, ]$, and, under Assumption~2, we have $ \mathbb{E} \, [ \, Y \mid X = x,  T = t \, ] = $ $\mathbb{E} \, [ \, Y \mid X = x, T = t, R = 1 \, ] $. Hence, in our setting, we can unbiasedly estimate CATE by learning a function $f_t : \mathcal{X} \rightarrow \mathcal{Y}$ for $t = 0,1$, such that $f_t(x)$ approximates $\mathbb{E} \, [ \, Y  \mid  X = x, \, T = t, \, R = 1 \, ] $ for which we have fully observed data. Then, we have the CATE estimator given by 
\begin{equation}
\hat{\tau}(x) = f_1(x) - f_0(x).
\end{equation}


Learning $f_t(x)$ for $t = 0,1$ from data is a standard machine learning problem for which various methods can be used. However, while the above assumptions ensure unbiased estimation of $f_0(x)$ and $f_1(x)$ from data, the estimators can have high variance when the covariate distributions between treatment groups ($T=0$ and $T=1$) and/or between treatment missingness groups ($R=0$ and $R=1$) differ. To illustrate this with an example, consider a job training program (treatment $T$) offered to high- and low-skilled workers (covariate $X$). Let us assume that low-skilled workers rarely decide to participate in the program (\ie, $T=0$ predominantly) and also rarely share the information about their participation (\ie, $R=0$ predominantly). In this case, we can have a high error when estimating $f_0(x)$ and $f_1(x)$ for low-skilled workers due to the lack of observed treatment data for this group. Moreover, we can have an even higher error when estimating $f_1(x)$ since not many low-skilled workers participated in the job training program (\ie, even when we observe the treatment for low-skilled workers, we have $T=0$ predominantly).


Hence, in the presence of different covariate distributions, standard methods may give unreliable CATE estimates due to high estimation variance in the covariate domains where observed data are lacking. The problem is that we fully observe data only from distribution $p(x,t,R=1)$ (\ie, the factual domain with observed treatment), but reliable CATE estimation also requires accurate outcome predictions in the missing treatment domain ($p(x,t,R=0)$), as well as in the counterfactual domain ($p(x,1-t,r)$). However, for both, we do not have fully observed data (\ie, we have missing treatment information and missing counterfactual outcomes, respectively). By observing that $p(x,t,r) = p(x) \, p(t \mid x) \, p(r \mid x)$ (under the causal structure in Fig.~\hyperref[fig:causal_structure]{1}), we see that the differences in the covariate distributions between these domains come from distributional differences (i)~between $p(t \mid x)$ and $p(1-t \mid x)$, and (ii)~between $p(R=1 \mid x)$ and $p(R=0 \mid x)$. We frame these distributional differences as covariate shifts. 


Therefore, we identify two covariate shifts in our setting with missing treatments: (i)~a covariate shift between the treated and control population, and (ii)~a covariate shift between the observed and missing treatment population. These covariate shifts could lead to high CATE estimation errors in covariate domains where data are not fully observed. In this paper, we develop a novel CATE estimation algorithm which addresses these covariate shifts and thus provides more reliable CATE estimates by reducing the estimation error. In the following section, we first mathematically show the effect of these covariate shifts by deriving a generalization bound for CATE estimation in our setting with missing treatments. The bound then serves as a theoretical foundation for our proposed algorithm.


\section{THEORY: GENERALIZATION BOUND} 
\label{sec:theory}


Our intuition from the previous section suggests that the expected error of CATE estimation depends on three error sources: (i)~the standard estimation error; (ii)~the covariate shift between the treated and control population; and (iii)~the covariate shift between the observed and missing treatment population. Here, we mathematically underpin this intuition and, to this end, derive a generalization bound in three steps:
\begin{itemize}
\item\emph{Step 1.} We bound the overall loss with the sum of the factual loss and the counterfactual loss (Lemma~1).
\item\emph{Step 2.} We bound the factual and counterfactual loss in the missing treatment domain using the corresponding losses in the observed treatment domain and the distance between the covariate distributions of the observed and missing treatment population (Lemma~2).
\item\emph{Step 3.} We bound the counterfactual loss in the observed treatment domain using the corresponding factual loss and the distance between the covariate distributions of the treated and control population (Lemma~3).
\end{itemize}
The lemmas then imply our main theoretical result provided in Theorem~1: the expected error of CATE estimation with missing treatments is bounded by the sum of (i)~the factual loss in the observed treatment domain (\ie, the standard generalization error); (ii)~the covariate distribution distance between the treated and control population; and (iii)~the covariate distribution distance between the observed and missing treatment population. The proofs and further details on theoretical results are in Supplement~\hyperref[app:proof]{A}.


In order to derive the generalization bound for CATE estimation, we define the (overall) estimation error in our setting. The standard CATE estimation error is given by the expected precision in estimation of heterogeneous effect~(PEHE) \citep{Hill2011}, which is basically the mean squared error of estimating $\tau(x)$. We adjust the PEHE for our setting with missing treatments and define the PEHE loss of a function $f$ as
\begin{equation}
\epsilon_{\mathrm{PEHE}}(f) = \int_{\mathcal{X} \times \mathcal{R}} \left( \hat{\tau}(x) - \tau(x) \right)^2 \, p(x,r) \, \mathrm{d}x \, \mathrm{d}r.
\end{equation}


We consider $f_t = h_t \circ \Phi$ for $t=0,1$, where $\Phi: \mathcal{X} \rightarrow \mathcal{Z}$ is a representation function, and $h_t : \mathcal{Z} \rightarrow \mathcal{Y}$ is a hypothesis defined over the representation space $\mathcal{Z}$. Hence, we have $f_t(x) = h_t(\Phi(x))$. We further use $f$ and the pair $(\Phi, h)$ interchangeably. We assume that the representation $\Phi : \mathcal{X} \rightarrow \mathcal{Z}$ is a one-to-one function and define $\Psi: \mathcal{Z} \rightarrow \mathcal{X}$ to be the inverse of $\Phi$, such that $\Psi(\Phi(x)) = x$ for all $x \in \mathcal{X}$. Moreover, by mapping the covariate space $\mathcal{X}$ with distribution $p$ onto the representation space $\mathcal{Z}$, the representation $\Phi$ induces a corresponding distribution $p_{\Phi}$ over $\mathcal{Z}$. 


\underline{Step 1}. In the first step, we bound the overall PEHE loss $\epsilon_{\mathrm{PEHE}}(f)$ with a sum of losses in the factual and counterfactual domain. Let $L_Y: \mathcal{Y} \times \mathcal{Y} \rightarrow \mathbb{R}_{+}$ be a loss function (\eg, squared loss). Then, we define the expected loss of $\Phi$ and $h$ for a covariates-treatment pair $(x,t)$ as
\begin{equation}
l_{h,\Phi}(x,t) = \int_{\mathcal{Y}} L_Y(y_t, h_t(\Phi(x))) \, p(Y_t=y_t \mid x) \, \mathrm{d} y_t.
\end{equation}
Note that the expected loss $l_{h,\Phi}(x,t)$ for a given pair $(x,t)$ does not depend on treatment missingness, since $R$ is conditionally independent of $Y_t$ given $X$. The expected factual and counterfactual losses of $\Phi$ and $h$ are given by
\begin{align}
\epsilon_{\mathrm{F}}(h, \Phi) &= \int_{\mathcal{X} \times \mathcal{R} \times \mathcal{T}} \!\!\!\!\!\!\!\!\!\!\!\!\!\!\! l_{h,\Phi}(x,t) \, p(x,r,t) \, \mathrm{d}x \, \mathrm{d}r \, \mathrm{d}t, \\
\epsilon_{\mathrm{CF}}(h, \Phi) &= \int_{\mathcal{X} \times \mathcal{R} \times \mathcal{T}} \!\!\!\!\!\!\!\!\!\!\!\!\!\!\! l_{h,\Phi}(x,t) \, p(x,r,1-t) \, \mathrm{d}x \, \mathrm{d}r \, \mathrm{d}t.
\end{align}


\textbf{Lemma 1} \emph{Let $\Phi: \mathcal{X} \rightarrow \mathcal{Z}$ be an invertible representation function and $h_t: \mathcal{Z} \rightarrow \mathcal{Y}$ for $t=0,1$ a hypothesis. Let $L_Y: \mathcal{Y} \times \mathcal{Y} \rightarrow \mathbb{R}_{+}$ be the squared loss. Then, we have}
\begin{equation}
\label{eq:lemma1}
\epsilon_{\mathrm{PEHE}}(h, \Phi) \leq 2 \, \left( \epsilon_{\mathrm{F}}(h, \Phi) + \epsilon_{\mathrm{CF}}(h, \Phi) - 4\sigma_{Y}^2 \right),
\end{equation}
\emph{where $\sigma_{Y}^2$ is the minimal variance of potential outcomes as defined in Definition~8 of Supplement~\hyperref[app:proof]{A}.}  


Lemma~1 provides a bound on $\epsilon_{\mathrm{PEHE}}(\Phi,h)$ using the sum of the factual and counterfactual loss, \ie, $\epsilon_{\mathrm{F}}(h, \Phi)$ and $\epsilon_{\mathrm{CF}}(h, \Phi)$. However, the problem is that, for $R=0$, we neither can estimate $\epsilon_{\mathrm{F}}(h, \Phi)$ and $\epsilon_{\mathrm{CF}}(h, \Phi)$ from data due to missing treatment information nor can we estimate $\epsilon_{\mathrm{CF}}(h, \Phi)$ in general due to missing counterfactual outcomes. Here, our idea is to bound these inestimable terms using their estimable counterparts and corresponding distributional distances induced by the representation. Hence, in Step~2, we bound the factual and counterfactual loss in the missing treatment domain using the corresponding losses in the observed treatment domain and the distance between the observed and missing treatment population (Lemma~2). Then, in Step~3, we bound the counterfactual loss in the observed treatment domain using the factual loss in the observed treatment domain and the distance between the treated and control population (Lemma~3). The three lemmas then directly imply our final bound (Theorem~1). 


\underline{Step 2}. We first introduce notation for the corresponding factual and counterfactual loss in the observed and missing treatment domain. We also define a distributional distance metric. We use superscripts to denote when we condition on a given variable, \eg, $p^{R=0}(x) = p(x \mid R=0)$. Then, the expected factual and counterfactual losses of $\Phi$ and $h$ in the domain $R=r$ for $r=0,1,$ (\ie, missing and observed treatment domain) are given by
\begin{align}
\epsilon_{\mathrm{F}}^{R=r}(h, \Phi) &= \int_{\mathcal{X} \times \mathcal{T}} l_{h,\Phi}(x,t) \, p^{R=r}(x,t) \, \mathrm{d}x \, \mathrm{d}t, \\
\epsilon_{\mathrm{CF}}^{R=r}(h, \Phi) &= \int_{\mathcal{X} \times \mathcal{T}} l_{h,\Phi}(x,t) \, p^{R=r}(x,1-t) \, \mathrm{d}x \, \mathrm{d}t. 
\end{align}
To measure distributional distances, we use the integral probability metric~(IPM), which is a class of metrics between probability distributions \citep{Muller1997, Sriperumbudur2012}. Let $G$ be a function family consisting of functions $g: \mathcal{S} \rightarrow \mathbb{R}$. For a pair of distributions $p_1,p_2$ over some space $\mathcal{S}$, the IPM is defined by
\begin{equation}
\mathrm{IPM}_G(p_1,p_2) = \sup_{g \in G} \bigg| \int_{\mathcal{S}} g(s) \, (p_1(s) - p_2(s)) \, \mathrm{d}s \, \bigg|.
\end{equation}
Thus, $\mathrm{IPM}_G(\cdot,\cdot)$ is a pseudo-metric on the space of probability functions over $\mathcal{S}$. For a sufficiently rich function family $G$, $\mathrm{IPM}_G(\cdot,\cdot)$ is a true metric over the corresponding set of probabilities, \ie, $\mathrm{IPM}_G(p_1,p_2) = 0 \Rightarrow p_1 = p_2$. 


\textbf{Lemma 2} \emph{Let $\Phi: \mathcal{X} \rightarrow \mathcal{Z}$ be an invertible representation and $\Psi$ its inverse. Let $p_{\Phi}$ be the distribution induced by $\Phi$ over $\mathcal{Z}$. Let $v = p(R=0)$. Let $G$ be a family of functions $k: \mathcal{Z} \rightarrow \mathbb{R}$ and $\mathrm{IPM}_G(\cdot,\cdot)$ the integral probability metric induced by $G$. Let $h_t: \mathcal{Z} \rightarrow \mathcal{Y}$ for $t=0,1$ be a hypothesis. Assume there exists a constant $B_{\Phi} > 0$, such that, for $t=0,1$, the function $g_{\Phi,h}(z) := \frac{1}{B_{\Phi}} l_{h,\Phi}(\Psi(z),t) \in G$. Then, we have}
\begin{align}
\label{eq:lemma2}
& \epsilon_{\mathrm{F}}(h, \Phi) + \epsilon_{\mathrm{CF}}(h, \Phi)   \nonumber  \\
\leq \, & \epsilon_{\mathrm{F}}^{R=1}(h, \Phi) + \epsilon_{\mathrm{CF}}^{R=1}(h, \Phi) \,\,  \\
& + 2 \, v \,  B_{\Phi} \, \mathrm{IPM}_G\big(p_{\Phi}^{R=0}(z), p_{\Phi}^{R=1}(z)\big) \nonumber.
\end{align}

\underline{Step 3}. The remaining inestimable term following Lemma~2 is the counterfactual loss in the observed treatment domain. However, we cannot estimate it due to missing counterfactual outcomes. Hence, in Lemma~3, we bound this term as well. 


\textbf{Lemma 3} \emph{Let $\Phi: \mathcal{X} \rightarrow \mathcal{Z}$ be an invertible representation and $\Psi$ its inverse. Let $p_{\Phi}$ be the distribution induced by $\Phi$ over $\mathcal{Z}$. Let $u = p(T=0)$. Let $G$ be a family of functions $g: \mathcal{Z} \rightarrow \mathbb{R}$ and $\mathrm{IPM}_G(\cdot,\cdot)$ the integral probability metric induced by $G$. Let $h_t: \mathcal{Z} \rightarrow \mathcal{Y}$ for $t=0,1$ be a hypothesis. Assume there exists a constant $B_{\Phi} > 0$, such that, for $t=0,1$, the function $g_{\Phi,h}(z) := \frac{1}{B_{\Phi}} l_{h,\Phi}(\Psi(z),t) \in G$. Then, we have}
\begin{align}
\label{eq:lemma3}
& \epsilon_{\mathrm{CF}}^{R=1}(h, \Phi)  \nonumber \\
\leq \, &  u \,  \epsilon_{\mathrm{F}}^{R=1, T=1}(h, \Phi) + (1-u) \,  \epsilon_{\mathrm{F}}^{R=1, T=0}(h, \Phi) \, \,   \\
& + B_{\Phi} \, \mathrm{IPM}_G\big(p_{\Phi}^{R=1,T=0}(z), p_{\Phi}^{R=1,T=1}(z)\big) \nonumber.
\end{align}

Given the above lemmas, we state the generalization bound as the main result of our paper in Theorem~1.  


\textbf{Theorem 1} \emph{Let $\Phi: \mathcal{X} \rightarrow \mathcal{Z}$ be an invertible representation and $\Psi$ its inverse. Let $p_{\Phi}$ be the distribution induced by $\Phi$ over $\mathcal{Z}$. Let $v = p(R=0)$. Let $G$ be a family of functions $g: \mathcal{Z} \rightarrow \mathbb{R}$ and $\mathrm{IPM}_G(\cdot,\cdot)$ the integral probability metric induced by $G$. Let $h_t: \mathcal{Z} \rightarrow \mathcal{Y}$ for $t=0,1$ be a hypothesis. Let $L_Y: \mathcal{Y} \times \mathcal{Y} \rightarrow \mathbb{R}_{+}$ be the squared loss function. Assume there exists a constant $B_{\Phi} > 0$, such that, for $t=0,1$, the function $g_{\Phi,h}(z) := \frac{1}{B_{\Phi}} l_{h,\Phi}(\Psi(z),t) \in G$. Then, we have}
\begin{align}
\label{eq:theorem}
&\epsilon_{\mathrm{PEHE}}(h, \Phi) \nonumber \\
 \leq \, & 2 \, \bigg[ \epsilon_{\mathrm{F}}^{R=1, T=1}(h, \Phi) + \epsilon_{\mathrm{F}}^{R=1, T=0}(h, \Phi) \,\,   \\
&  \quad + B_{\Phi} \, \mathrm{IPM}_G(p_{\Phi}^{R=1, T=0}(z), p_{\Phi}^{R=1, T=1}(z))  \nonumber \\
& \quad + 2 \, v \,  B_{\Phi} \, \mathrm{IPM}_G\big(p_{\Phi}^{R=0}(z), p_{\Phi}^{R=1}(z)\big) - 4\sigma_{Y}^2 \bigg] \nonumber.
\end{align}


Theorem~1 shows that the expected CATE estimation error $\epsilon_{\mathrm{PEHE}}(h, \Phi)$ for a representation $\Phi$ and hypothesis $h$ is bounded by a sum of (i)~the standard generalization error for that representation ($\epsilon_{\mathrm{F}}^{R=1, T=1}(h, \Phi) + \epsilon_{\mathrm{F}}^{R=1, T=0}(h, \Phi)$); (ii)~the distance between the treated and control distributions induced by the representation ($\mathrm{IPM}_G(p_{\Phi}^{R=1, T=0}(z), p_{\Phi}^{R=1, T=1}(z))$); and (iii)~the distance between the observed and missing treatment distributions induced by the representation ($\mathrm{IPM}_G(p_{\Phi}^{R=0}(z), p_{\Phi}^{R=1}(z))$). The bound shows different sources of error when estimating CATE with missing treatments, \ie, the standard generalization error and the two covariate shifts formalized using the IPM metric. 


We make a few additional remarks regarding the derived generalization bound. The IPM terms reflect the two described covariate shifts. Both evaluate to zero in case that the covariate distributions are balanced with respect to treatment and treatment missingness, \ie, when covariates $X$ neither affect treatment $T$ nor treatment missingness $R$. The IPM term that reflects the distribution imbalance with respect to treatment missingness (\ie, $\mathrm{IPM}_G(p_{\Phi}^{R=0}(z), p_{\Phi}^{R=1}(z))$) is scaled by the probability of missingness $v = p(R=0)$, meaning that its relative importance depends on $v$. In other words, when we have a small probability of treatment missingness, the corresponding covariate shift between the observed and missing treatment population is relatively less important compared to the other two sources of error. Moreover, when the probability of missingness, $v$, equals zero, our generalization bound reduces to the generalization bound for CATE estimation in the standard setting \citep{Shalit2017}. Hence, we provide a \emph{different} bound in a more general setting.



The derived generalization bound holds for any given invertible representation $\Phi$ and hypothesis $h$ that satisfy the conditions of Theorem~1. Given empirical data and representation-hypothesis space, we can upper bound the loss terms $\epsilon_{\mathrm{F}}^{R=1, T=1}(h, \Phi)$ and $\epsilon_{\mathrm{F}}^{R=1, T=0}(h, \Phi)$ with their empirical counterparts and model complexity terms by applying standard machine learning theory \citep{Shalev2014}. This naturally leads to a CATE estimation algorithm based on representation learning that minimizes the upper bound in Eq.~(\ref{eq:theorem}): (i)~by minimizing the empirical version of the loss terms $\epsilon_{\mathrm{F}}^{R=1, T=1}(h, \Phi)$ and $\epsilon_{\mathrm{F}}^{R=1, T=0}(h, \Phi)$, and (ii)~by minimizing respective IPM terms using either the empirical IPM distances as in \citet{Shalit2017} or via adversarial learning \citep{Ganin2016}. Here, we use adversarial learning.

\section{CATE ESTIMATION ALGORITHM}
\label{sec:algo}

\begin{figure}[tbp]
\centerline{\includegraphics[width= 6cm]{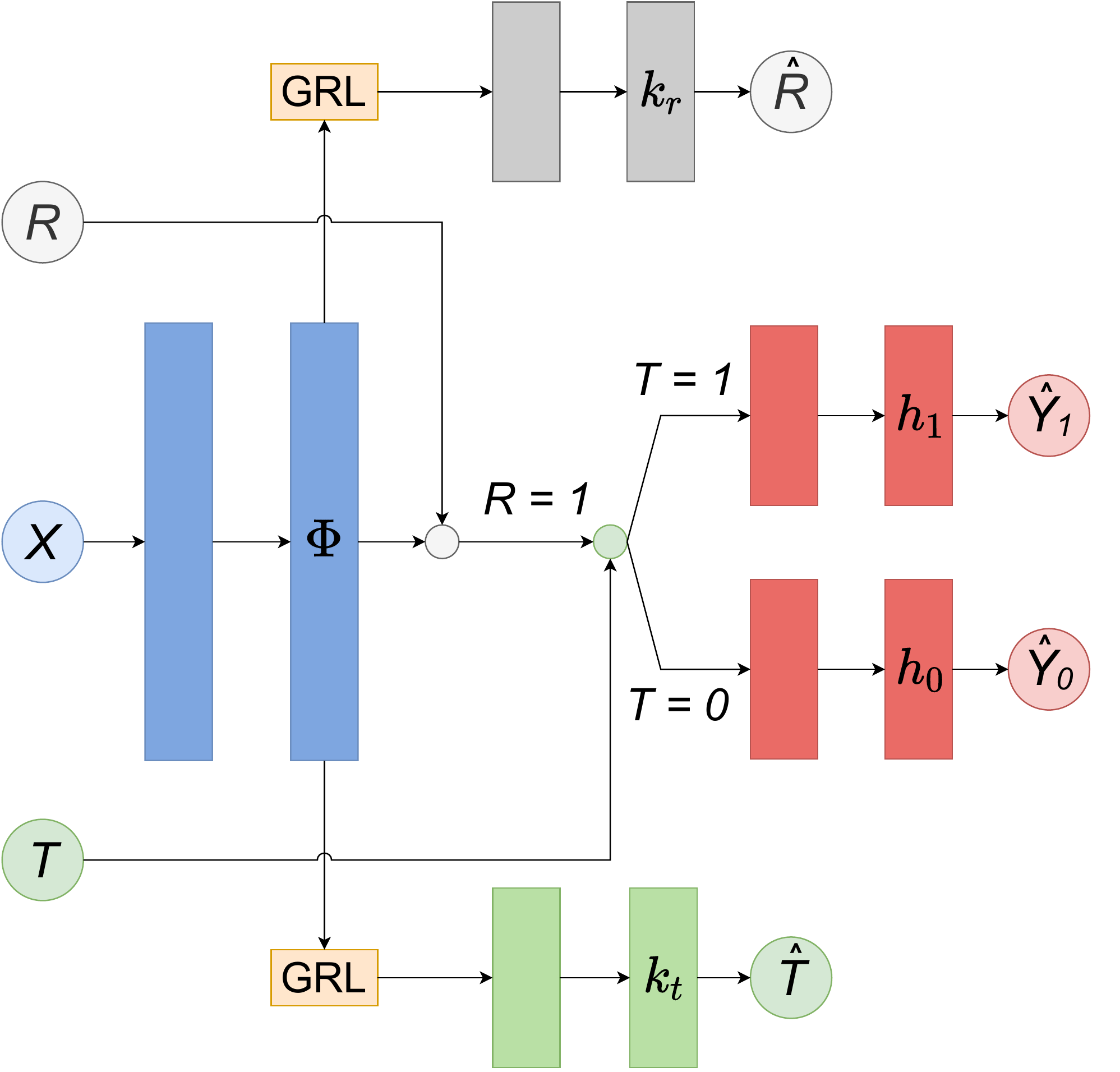}}
\caption{Overview of our MTRNet.}
\label{fig:mthd_arch}
\end{figure}

In this section, we propose the \emph{missing treatment representation network}~(MTRNet), our algorithm for CATE estimation in the setting with missing treatment information. The architecture of MTRNet is shown in Fig.~\hyperref[fig:mthd_arch]{2}. For given data $\mathcal{D} =  \{(t_i, r_i, x_i, y_i)\}_{i=1}^{n} $, MTRNet minimizes a novel empirical loss based on our generalization bound from Theorem~1. The corresponding objective function is given by
\begin{align}
\label{eq:obj}
\min_{\substack{\Phi, h \\ \| \Phi \| = 1}} & \frac{1}{n_o} \sum \limits_{\forall i: r_i=1} w_i \, L_Y(h_{t_i}(\Phi(x_i)),y_i) + \lambda \, \| W_h \|_{2}^2     \nonumber \\
 & -  \, \, \alpha \, \frac{1}{n_o} \sum \limits_{\forall i: r_i=1} L_T(k_t(\Phi(x_i)),t_i)  \\  
& - \, \, \beta  \frac{1}{n}  \sum \limits_{i=1}^n L_R(k_r(\Phi(x_i)),r_i) \nonumber, 
\end{align}
\mbox{with $w_i = \frac{t_i}{2u} + \frac{1-t_i}{2 \, (1-u)}$, $u = \frac{1}{n_o} \sum \limits_{\forall i: r_i=1} t_i$, and $n_o = \sum \limits_{i=1}^n r_i $}. 


In Eq.~(\ref{eq:obj}), we replaced the theoretical loss terms from the bound by their corresponding empirical ones. The standard generalization error, \ie, $\epsilon_{\mathrm{F}}^{R=1, T=1}(h, \Phi) + \epsilon_{\mathrm{F}}^{R=1, T=0}(h, \Phi)$, is replaced by a weighted outcome prediction loss, $ w_i \, L_Y(h_{t_i}(\Phi(x_i)),y_i)$, where the weights reflect the size of the treated and control population. The IPM terms, \ie, $\mathrm{IPM}_G\big(p_{\Phi}^{R=1, T=0}(z), p_{\Phi}^{R=1, T=1}(z)\big)$ and $\mathrm{IPM}_G\big(p_{\Phi}^{R=0}(z), p_{\Phi}^{R=1}(z)\big)$, are minimized by adding a negative prediction loss for treatment (\ie, $ L_T(k_t(\Phi(x_i)),t_i) $ with prediction function $k_t$), as well as for treatment missingness (\ie, $  L_R(k_r(\Phi(x_i)),r_i)$ with prediction function $k_r$), respectively. The rationale for minimizing the IPM terms through maximizing the prediction losses is the following: For a covariate representation $\Phi$, when prediction loss $L_R$ is large, the representation is not predictive of $R$, and, hence, it is not informative of whether a data point belongs to the group $R=0$ or the group $R=1$. Consequently, the distribution $p_{\Phi}$ induced by the representation for the group $R=0$ is similar to the one for the group $R=1$, which means that the corresponding IPM term, \ie, $\mathrm{IPM}_G\big(p_{\Phi}^{R=0}(z), p_{\Phi}^{R=1}(z)\big)$, is small. Hence, maximizing $L_R$ (\ie, minimizing $-L_R$) aims to minimize the corresponding IPM term. The rationale for $T$ is analogous. We maximize these prediction losses for a representation $\Phi$ using adversarial learning with gradient reversal layer (GRL). The GRL reverses the gradient for representation layers during learning such that the learned representation aims to maximize the prediction loss instead of minimizing it \citep{Ganin2016}. Since constant $B_{\Phi}$ cannot be evaluated for a general function family \citep{Shalit2017}, we use hyperparameters $\alpha$ and $\beta$ to trade-off outcome prediction accuracy and reducing the respective IPM distances. We also introduce an $L_2$-regularization with parameter $\lambda$ for the weights of the hypothesis layers $W_h$, and use batch normalization to fix the norm of $\Phi$. 



\begin{algorithm}[tbp]
\caption{Learning algorithm for MTRNet}
\label{alg:MTRNet}
\scriptsize
\begin{algorithmic}[1]
\scriptsize
\Require{Data $\mathcal{D} =  \{(t_i, r_i, x_i, y_i)\}_{i=1}^{n} $; loss functions: $L_Y$, $L_T$ and $L_R$; hyperparameters  $b$, $\eta$, $\lambda$, $\alpha$, $\beta$; and network architecture with weights $W_{\Phi}$, $W_h$, $W_{k_t}$, $W_{k_r}$}
\Ensure{Optimal representation $\Phi^*$ and hypothesis $h^*$ with weights $W_{\Phi}^*$ and $W_h^*$}
\Repeat
\State Randomly sample mini-batch of size $b$ from  $\mathcal{D}$
\State \mbox{Compute $n_o$, $u$, and  $w_i$ for $i=1, \ldots,b$ using Eq.~(\ref{eq:obj})}
\State \mbox{Compute $g_1 = \nabla_{W_{\Phi}} \frac{1}{n_o} \sum \limits_{\forall i: r_i=1} w_i \, L_Y(h_{t_i}(\Phi(x_i)),y_i) $}
\State \mbox{Compute $g_2 = \nabla_{W_{\Phi}} \frac{1}{n_o} \sum \limits_{\forall i: r_i=1} L_T(k_t(\Phi(x_i)),t_i) $}
\State \mbox{Compute $g_3 = \nabla_{W_{\Phi}}\frac{1}{b}  \sum \limits_{i=1}^b L_R(k_r(\Phi(x_i)),r_i)$}
\State \mbox{Compute $g_4 = \nabla_{W_h} \frac{1}{n_o} \sum \limits_{\forall i: r_i=1} w_i \, L_Y(h_{t_i}(\Phi(x_i)),y_i) $ }
\State \mbox{Compute $g_5 = \nabla_{W_{k_t}} \frac{1}{n_o} \sum \limits_{\forall i: r_i=1} L_T(k_t(\Phi(x_i)),t_i) $}
\State \mbox{Compute $g_6 = \nabla_{W_{k_r}} \ \frac{1}{b}  \sum \limits_{i=1}^b L_R(k_r(\Phi(x_i)),r_i)$}
\State Update weights 
\begin{align*}
W_{\Phi} &\leftarrow W_{\Phi} - \eta \, (g_1 + \alpha g_2 + \beta g_3) , \\
W_h &\leftarrow W_h - \eta \, (g_4 + 2\lambda W_h) , \\
W_{k_t} &\leftarrow  W_k + \eta \alpha g_5 ,  \\
W_{k_r} &\leftarrow W_r + \eta \beta g_6  
\end{align*}
\Until convergence
\end{algorithmic}
\end{algorithm}


MTRNet outputs the optimal $\Phi$ and $h$ based on the above objective. The learning algorithm is given in Algorithm~\ref{alg:MTRNet}. To train MTRNet, we use Adam \citep{Kingma2015} and run Algorithm~\ref{alg:MTRNet} for a given number of iterations. The network architecture of MTRNet comprises three representation layers for representation $\Phi$, three hypothesis layers for hypothesis $h_t$, for each $t=0,1$, and one layer for each prediction function, $k_t$ and $k_r$. We use exponential linear unit (ELU) \citep{Clevert2015} activation function with dropout. The hyperparameters include: representation layer size, hypothesis layer size, number of iterations, batch size, learning rate, dropout rate, $\lambda$, $\alpha$, and $\beta$. We choose hyperparameters via cross-validation with a $70/20/10$ split. Further implementation details are given in Supplement~\hyperref[app:exp_det]{B}.   

\section{EXPERIMENTS}
\label{sec:exp}

In this section, we show the effectiveness of our MTRNet for CATE estimation with missing treatments and, to do so, we use both semi-synthetic and real-world data. To this end, we demonstrate that, by addressing the covariate shifts, MTRNet reduces CATE estimation error across different covariate domains and thus provides superior overall performance compared to baseline methods.  


\begin{table*}[t]
\begin{center}
\resizebox{.95\textwidth}{!}{
\begin{tabular}{!{\vrule width 1pt} l !{\vrule width 1pt}c | c | c !{\vrule width 1pt} c | c | c !{\vrule width 1pt} c | c | c !{\vrule width 1pt}}
\toprule
\textbf{Method} &  \multicolumn{3}{c !{\vrule width 1pt}}{\textbf{IHDP} ($\sqrt{\hat{\epsilon}_{\mathrm{PEHE}}}$)} &  \multicolumn{3}{c !{\vrule width 1pt}}{\textbf{Twins} ($\sqrt{\bar{\epsilon}_{\mathrm{PEHE}}}$)} &  \multicolumn{3}{c !{\vrule width 1pt}}{\textbf{Jobs} ($\hat{R}_{\mathrm{Pol}}(\pi_f)$)} \\
\hline
  & $\mathrm{Overall}$ & $\mathrm{T_{observed}}$  & $\mathrm{T_{missing}}$ & $\mathrm{Overall}$ & $\mathrm{T_{observed}}$  & $\mathrm{T_{missing}}$ & $\mathrm{Overall}$ & $\mathrm{T_{observed}}$  & $\mathrm{T_{missing}}$ \\
\hline
$\mathrm{OLS_{del}}$ & $1.21 \pm .41$ & $1.14 \pm .33$  & $1.25 \pm .50$ & $.29 \pm .00$ & $.26 \pm .00$ & $.32 \pm .00$ & $.35 \pm .00$ & $.36 \pm .00$  & $.42 \pm .00$   \\
\hline
$\mathrm{OLS_{imp}}$ & $1.62 \pm .39$ & $1.49 \pm .36$  & $1.75 \pm .48$ & $.29 \pm .00$ & $.26 \pm .00$ & $.32 \pm .00$ & $.35 \pm .00$ & $.34 \pm .00$  & $.50 \pm .00$   \\
\hline
$\mathrm{OLS_{rew}}$ & $1.24 \pm .38$ & $1.19 \pm .30$  & $1.28 \pm .48$ & $.29 \pm .00$ & $.26 \pm .00$ & $.32 \pm .00$ & $.37 \pm .00$ & $.36 \pm .00$  & $.50 \pm .00$   \\
\hline
$\mathrm{CF_{del}}$ & $1.53 \pm .41$ & $1.51 \pm .42$  & $1.52 \pm .47$ & $.29 \pm .00$ & $.26 \pm .00$ & $.32 \pm .00$ & $.32 \pm .01$ & $.32 \pm .02$  & $.42 \pm .05$   \\
\hline
$\mathrm{CF_{imp}}$ & $1.68 \pm .53$ & $1.64 \pm .50$  & $1.71 \pm .59$ & $.29 \pm .00$ & $.26 \pm .00$ & $.32 \pm .00$ & $.32 \pm .02$ & $.31 \pm .03$  & $.42 \pm .00$   \\
\hline
$\mathrm{CF_{rew}}$ & $1.51 \pm .42$ & $1.49 \pm .42$  & $1.51 \pm .48$ & $.29 \pm .00$ & $.26 \pm .00$ & $.32 \pm .00$ & $.32 \pm .02$ & $.31 \pm .03$  & $.41 \pm .03$   \\
\hline
$\mathrm{TARNet_{del}}$ & $1.19 \pm .21$ & $1.26 \pm .24$  & $1.11 \pm .18$ & $.29 \pm .00$ & $.26 \pm .00$ & $.32 \pm .00$ & $.27 \pm .02$ & $.26 \pm .02$  & $.44 \pm .07$   \\
\hline
$\mathrm{TARNet_{imp}}$ & $1.76 \pm .54$ & $1.54 \pm .38$  & $1.94 \pm .80$ & $.29 \pm .00$ & $.26 \pm .00$ & $.32 \pm .00$ & $.27 \pm .02$ & $.26 \pm .02$  & $.45 \pm .08$   \\
\hline
$\mathrm{TARNet_{rew}}$ & $1.15 \pm .11$ & $1.16 \pm .17$  & $1.13 \pm .17$ & $.29 \pm .00$ & $.26 \pm .00$ & $.32 \pm .00$ & $.26 \pm .02$ & $.25 \pm .02$  & $.42 \pm .08$   \\
\hline
$\mathrm{CFRMMD_{del}}$ & $1.22 \pm .23$ & $1.25 \pm .23$  & $1.17 \pm .33$ & $.29 \pm .00$ & $.26 \pm .00$ & $.32 \pm .00$ & $.32 \pm .03$ & $.31 \pm .03$  & $.44 \pm .06$   \\
\hline
$\mathrm{CFRMMD_{imp}}$ & $1.50 \pm .31$ & $1.41 \pm .29$  & $1.58 \pm .42$ & $.30 \pm .01$ & $.27 \pm .01$ & $.33 \pm .01$ & $.27 \pm .02$ & $.26 \pm .03 $  & $.38 \pm .04$   \\
\hline
$\mathrm{CFRMMD_{rew}}$ & $1.29 \pm .32$ & $1.31 \pm .27$  & $1.27 \pm .40$ & $.29 \pm .00$ & $.26 \pm .00$ & $.32 \pm .00$ & $.32 \pm .03$ & $.32 \pm .03$  & $.41 \pm .06$   \\
\hline
\textbf{MTRNet} (ours) & $\mathbf{1.00 \pm .23}$ & $\mathbf{1.03 \pm .25}$  & $\mathbf{0.96 \pm .28}$ & $\mathbf{.28 \pm .00}$ & $\mathbf{.26 \pm .00}$ & $\mathbf{.31 \pm .00}$ & $\mathbf{.23 \pm .04}$ & $\mathbf{.24 \pm .05}$  & $\mathbf{.28 \pm .06}$ \\
\bottomrule 
\multicolumn{10}{l}{* Lower is better (best in bold).} \\
\end{tabular}}
\label{tab:results}
\caption{Results of experiments on three benchmark datasets (mean averaged over 10 runs $\pm$ standard deviation).}
\end{center}
\end{table*}

\textbf{Baselines.} CATE estimation with missing treatments has been overlooked by the existing literature. Hence, appropriate baselines are missing. Instead, we need to construct baselines by combining CATE estimation methods in the standard setting with different methods for dealing with missing data. Here, we use the following CATE estimation methods: (i)~linear model (\textbf{OLS}) fitted for each treatment group; (ii)~causal forest (\textbf{CF}) \citep{Athey2019}; (iii)~treatment agnostic representation network (\textbf{TARNet}) \citep{Shalit2017}; and (iv)~counterfactual regression maximum mean discrepancy (\textbf{CFRMMD}) \citep{Shalit2017}. Note that none of above methods address the covariate shift between the observed and missing treatment population since neither our setting nor this particular covariate shift were considered by the existing work. 


We combine the above methods with common methods for dealing with missing data \citep{Williamson2012}: (i)~deleting data points with missing treatment~(\textbf{del}); (ii)~imputing missing treatments using a machine learning model~(\textbf{imp}); and (iii)~re-weighting data points with observed treatment by the inverse probability of treatment being observed~(\textbf{rew}). In cases (i) and (ii), we first apply a missing data method to the initial data to deal with missing treatment information, and then, apply a CATE estimation method on the resulting complete data. In case (iii), we first estimate a model for the probability that the treatment is observed given covariates by using the initial data, and, then, apply a CATE estimation method on the complete part of the initial data (\ie, data with observed treatment), where each data point is weighted by the inverse of the estimated probability that the treatment is observed given its covariates. For imputation and re-weighting, we use random forest to model the respective probabilities. By combining the above CATE estimation methods with methods for dealing with missing data, we obtain 12 baselines in total. We name the baselines using the CATE method name and the method for dealing with missing data as subscript (\eg, $\mathrm{OLS_{del}}$ means OLS combined with deletion of data points with missing treatment).


\textbf{Datasets.} We conduct experiments with three benchmark datasets for CATE estimation but modify them such that treatment information is partially missing. Note that our method is directly applicable for CATE estimation from observational data with missing treatments in practice. However, available observational datasets with missing treatments cannot be used to evaluate the estimated CATE since the true CATE is unknown. Hence, we use the best practice for evaluating CATE estimation, and modify benchmark datasets for CATE estimation such that they fit to our setting with missing treatment information. The mechanism for introducing missingness is designed such that treatment missingness $R$ depends on covariates $X$ (as in our setting, see Fig.~\hyperref[fig:causal_structure]{1}). This way, we introduce both missing treatments and the covariate shift between the observed and missing treatment population. The proportion of data with missing treatment information is controlled by a parameter $m \in (0,1)$, and the magnitude of the covariate shift by a parameter $q \in (0,1)$. Details are in Supplement~\hyperref[app:exp_det]{B}. 

We use the following benchmark datasets: (i)~\textbf{IHDP} \citep{Hill2011, Shalit2017, hatt2021estimating}: a semi-synthetic dataset with covariates from a randomized experiment and outcomes simulated using a domain-specific probabilistic model. Hence, noiseless outcomes and the true CATE are available for this dataset. (ii)~\textbf{Twins} \citep{Almond2005, Yoon2018, hatt2021estimating}: a semi-synthetic dataset where the treatment assignment is simulated. Here, we do not observe the true CATE but we observe both potential outcomes. (iii)~\textbf{Jobs} \citep{LaLonde1986, Smith2005, Shalit2017}: real-world dataset that combines a randomized controlled trial~(RCT) and a larger observational dataset. Here, we do not have information about the true CATE; however, the randomized portion of the data still allows for evaluating CATE estimation error using policy risk (explained later).


\textbf{Performance metrics.} We evaluate the CATE estimation performance in different ways depending on the above datasets, \ie, depending on whether the true CATE is available. (i)~\textbf{IHDP}: we use the empirical PEHE given by $\hat{\epsilon}_{\mathrm{PEHE}} = \frac{1}{n} \sum_{i=1}^n  (\hat{\tau}(x) - \tau(x))^2$, thereby reflecting that we have access to the true CATE. (ii)~\textbf{Twins}: we use the observed PEHE given by $\bar{\epsilon}_{\mathrm{PEHE}} = \frac{1}{n} \sum_{i=1}^n  (\hat{\tau}(x) - (y_{1i} - y_{0i}) )^2 $ since we observe both potential outcomes, $Y_1$ and $Y_0$, but we cannot access information on the true CATE. (iii)~\textbf{Jobs}: we cannot evaluate the PEHE loss because we can neither access the true CATE nor the counterfactual outcomes. Instead, we use the policy risk that measures the average loss in value when treating according to the policy suggested by a CATE estimator. For a given model $f$, we define the policy $\pi_f(x)$ to be: treat $\pi_f(x) = 1$ if $\hat{\tau}(x) > 0$, and do not treat $\pi_f(x) = 0$ otherwise. Then, the policy risk is given by $R_{\mathrm{Pol}}(\pi_f) = 1 - \big(\mathbb{E}[Y_1 \mid \pi_f(x) = 1] \, p(\pi_f(x) = 1) +$ $\mathbb{E}[Y_0 \mid \pi_f(x) = 0] \, p(\pi_f(x) = 0)\big)$. Here, we compute the empirical policy risk $\hat{R}_{\mathrm{Pol}}(\pi_f)$ using the randomized portion of the data. 

\textbf{Results.} Table~\hyperref[tab:results]{1} shows the performance of our MTRNet vs. the 12 baselines for different experiments using the IHDP, Twins, and Jobs datasets. We report the mean performance averaged over 10 runs with the corresponding standard deviation. For each dataset, we report the overall error, the error in the observed treatment domain ($\mathrm{T_{observed}}$), and the error in the missing treatment domain ($\mathrm{T_{missing}}$). 

We make two important observations. (i)~MTRNet achieves the lowest overall error across all three datasets. This shows that our algorithm is effective for CATE estimation in the setting with missing treatments. On top of that, it provides superior CATE estimates compared to the state-of-the-art baselines. (ii)~The improvement in the overall CATE estimation by MTRNet comes from a substantially better performance in the missing treatment domain. Hence, by addressing the covariate shift between the observed and missing treatment population, MTRNet achieves a lower error when estimating CATE in the missing treatment domain (\ie, the covariate domain where CATE estimation is impeded due to the lack of fully observed data) compared to the baselines which ignore this covariate shift. This stresses the importance of addressing this aforementioned covariate shift in settings with missing treatment information. So far, this issue that has been overlooked by previous literature.

\begin{figure}[tbp]
\centerline{\includegraphics[width= .9\linewidth]{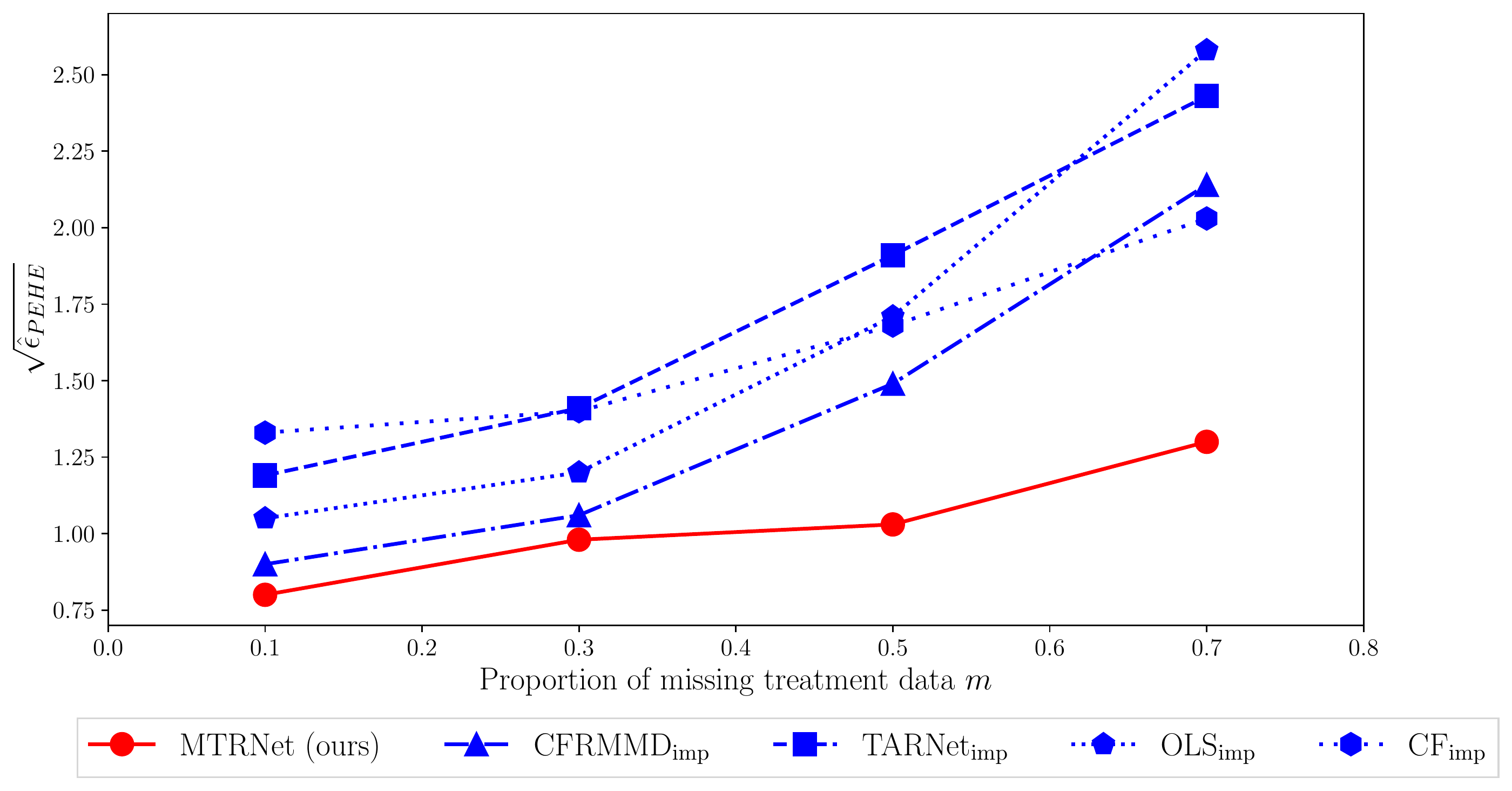}}
\caption{{IHDP results for varying parameter $m$.}}
\label{fig:m_results}
\end{figure}

The results in Table~\hyperref[tab:results]{1} were obtained in experiments where the proportion of missing treatment data was fixed to $m = 0.5$. In Fig.~\hyperref[fig:m_results]{3}, we show the results of IHDP experiments when varying parameter $m$. These results show the performance of MTRNet and the four CATE estimation methods combined with a method for imputing missing treatments (similar results with deletion and re-weighting are given in Supplement~\hyperref[app:add_exp]{C}). We see that, as we increase the proportion of data with missing treatment information ($m$), the performance gap between our MTRNet (in {\color{red}red}) and the baseline methods (in {\color{blue}blue}) becomes larger. This means that addressing the covariate shift between the observed and missing treatment population becomes more important, the higher is the probability that treatments are missing, which is also in line with our theoretical result in Theorem~1. Hence, addressing the covariate shift between the observed and missing treatment population is essential for reliable CATE estimation in settings with missing treatments, especially in case of large rates of missing treatments. 

\section{DISCUSSION}
\label{sec:disc}

In this paper, we analyzed CATE estimation in the setting with missing treatments, which, as shown above, presents unique challenges in the form of covariate shifts. Specifically, we identified two covariate shifts in our setting: (i)~a covariate shift between the treated and control population, and (ii)~a covariate shift between the observed and missing treatment population. While the covariate shift~(i) has been addressed in the existing CATE estimation literature, both the setting with missing treatments and the covariate shift~(ii) have been overlooked by the existing work. 

We fill this research gap from both theoretical and practical perspective. First we derived a generalization bound for CATE estimation with missing treatments that theoretically shows the effect of the two covariate shifts. Then, based on our bound, we proposed MTRNet, a novel CATE estimation algorithm that addresses these covariate shifts in our setting with missing treatments. We demonstrated that our MTRNet achieves superior performance in estimating CATE, especially in the missing treatment domain since it is the only CATE estimation algorithm that addresses the covariate shift between the observed and missing treatment population. The performance gain becomes even more pronounced when $m$, \ie, the treatment missingness rate, is large. The importance of our work is reflected by omnipresence of missing treatments in real-world applications. This holds true for both observational and RCT studies. Moreover, our MTRNet has direct practical implications as it provides more reliable CATE estimates that can improve personalized decision-making in many application areas, including personalized medicine.

\subsubsection*{Acknowledgements} 

We thank the reviewers for their valuable comments which allowed us to improve the paper. Funding through the Swiss National Science Foundation (186932) is acknowledged.

\bibliography{references}

\appendix

\onecolumn

\section{PROOF OF THEOREM 1}
\label{app:proof}

In our problem setup, we assume a distribution $ p(t, r, x, y_0, y_1)$ with the following variables: assigned treatment $T \in \mathcal{T} = \{0,1\} $, treatment missingness $R \in \mathcal{R} = \{0,1\} $, covariates $X \in \mathcal{X} = \mathbb{R}^d $, and potential outcomes $ Y_0, Y_1 \in \mathcal{Y} = \mathbb{R}$. We observe only one of the two potential outcomes, \ie, we observe $Y \in \mathcal{Y} = \mathbb{R}$, where $Y = Y_0$ or $Y = Y_1$, depending on the assigned treatment $T=t$. The observed potential outcome corresponding to the assigned treatment $t$ is called the \emph{factual} outcome, and the unobserved potential outcome corresponding to the other treatment possibility (\ie, $1-t$) is called the \emph{counterfactual} outcome. 

Our objective is to estimate the conditional average treatment effect~(CATE) for an individual with covariates $X=x$.

\textbf{Definition 1} \emph{The conditional average treatment effect~(CATE) for an individual with covariates $X=x$ is given by}
\begin{align*}
\tau(x) :&= \mathbb{E} \, [ \, Y_1 - Y_0 \mid X = x \, ].
\end{align*}

We make the following assumptions needed for identification of CATE in the setting with missing treatments:

\textbf{Assumption 1} \emph{(Consistency, $T$-Positivity, $T$-Ignorability)}.

(i) \, \quad \emph{$Y = Y_0$ if $T = 0$, and $Y = Y_1$ if $T = 1$ (Consistency);}

(ii) \ \quad \emph{$0 < p(T = 1  \mid  X = x) < 1$ if $p(x) \neq 0 $ ($T$-Positivity);}

(iii) \quad \emph{$Y_0, Y_1 \perp \!\!\! \perp T \mid X = x \, \, $ ($T$-Ignorability).}

\textbf{Assumption 2} \emph{($R$-Positivity, $R$-Ignorability)}.

(i) \, \quad \emph{$0 < p(R = 1  \mid  X = x) < 1$ if $p(x) \neq 0 $ ($R$-Positivity);}

(ii) \quad \emph{$ R \perp \!\!\! \perp T, Y_0, Y_1 \mid X = x \, \, $ ($R$-Ignorability).}

Under the above assumptions we have that $\mathbb{E} \, [ \, Y_t \mid X = x \, ] = \mathbb{E} \, [ \, Y \mid X = x,  T = t \, ] = $ $\mathbb{E} \, [ \, Y \mid X = x, T = t, R = 1 \, ] $. Hence, we can unbiasedly estimate CATE from data by learning a function $f_t : \mathcal{X} \rightarrow \mathcal{Y}$ for $t=0,1$. However, such estimation can have high variance in the presence of covariate shifts. 

In this work, we simultaneously address: (i)~the covariate shift between the observed and the missing treatment population, and (ii)~the covariate shift between the treated and the control population. We use a representation learning approach with $f_t = h_t \circ \Phi$, where $\Phi: \mathcal{X} \rightarrow \mathcal{Z}$ is a representation function, and $ h_t: \mathcal{Z} \rightarrow \mathcal{Y}$ for $t=0,1$ is a hypothesis defined over the representation space $\mathcal{Z}$. Hence, we have $f_t(x) = h_t(\Phi(x))$. Below, we define the estimator of CATE.

\textbf{Definition 2} \emph{The CATE estimator for an individual with covariates $X = x$ is given by}
\begin{equation*}
\hat{\tau}(x) = h_1(\Phi(x)) - h_0(\Phi(x)) = f_1(x) - f_0(x).
\end{equation*}

The estimation error for our setting with missing treatment information is given by the expected precision in estimation of heterogeneous effect (PEHE), \ie, the mean squared error in estimating $\tau(x)$.

\textbf{Definition 3} \emph{The PEHE loss of $\Phi$ and $h$ is given by}
\begin{equation*}
\epsilon_{\mathrm{PEHE}}(h, \Phi) = \int_{\mathcal{X} \times \mathcal{R}} \left(\hat{\tau}(x) - \tau(x) \right)^2 \, p(x,r) \, \mathrm{d}x \, \mathrm{d}r.
\end{equation*}

We make the following assumption about the representation function $\Phi$.

\textbf{Assumption 3} \emph{The representation $\Phi : \mathcal{X} \rightarrow \mathcal{Z}$ is a differentiable, invertible function. We assume that $\mathcal{Z}$ is the image of $\mathcal{X}$ under $\Phi$ and define $\Psi: \mathcal{Z} \rightarrow \mathcal{X}$ to be the inverse of $\Phi$, such that $\Psi(\Phi(x)) = x$ for all $x \in \mathcal{X}$.}

By mapping the covariate space $\mathcal{X}$ onto the representation space $\mathcal{R}$, the representation $\Phi$ induces a corresponding distribution $p_{\Phi}$.

\textbf{Definition 4} \emph{For a representation function $\Phi : \mathcal{X} \rightarrow \mathcal{Z}$ and for a distribution $p$ defined over $\mathcal{X}$, let $p_{\Phi}$ be the distribution induced by $\Phi$ over $\mathcal{Z}$}.

Let $L_Y: \mathcal{Y} \times \mathcal{Y} \rightarrow \mathbb{R}_{+}$ be a loss function, \eg, absolute or squared loss. 

\textbf{Definition 5} \emph{Let $\Phi : \mathcal{X} \rightarrow \mathcal{Z}$ be a representation function and $h_t: \mathcal{Z} \rightarrow \mathcal{Y}$ for $t=0,1$ a hypothesis defined over the representation space $\mathcal{Z}$. We define the expected loss for the covariates-treatment pair $(x,t)$ as}
\begin{equation*}
l_{h,\Phi}(x,t) = \int_{\mathcal{Y}} L_Y(y_t, h_t(\Phi(x))) \, p(Y_t = y_t \mid X=x) \, \mathrm{d}y_t.
\end{equation*}

Note that the expected loss $l_{h,\Phi}(x,t)$ for a given pair $(x,t)$ does not depend on treatment missingness, since we have conditional independence between $R$ and $Y_t$ given $X$. Next, we define losses in the factual and counterfactual domain, and the variance of $Y_t$ with respect to the distribution $p(x,r,t)$.

\textbf{Definition 6} \emph{The expected factual and counterfactual losses of $\Phi$ and $h$ are given by}
\begin{align*}
\epsilon_{\mathrm{F}}(h, \Phi) &= \int_{\mathcal{X} \times \mathcal{R} \times \mathcal{T}} \!\!\!\!\!\!\!\!\!\!\!\!\!\!\! l_{h,\Phi}(x,t) \, p(x,r,t) \, \mathrm{d}x \, \mathrm{d}r \, \mathrm{d}t, \\
\epsilon_{\mathrm{CF}}(h, \Phi) &= \int_{\mathcal{X} \times \mathcal{R} \times \mathcal{T}} \!\!\!\!\!\!\!\!\!\!\!\!\!\!\! l_{h,\Phi}(x,t) \, p(x,r,1-t) \, \mathrm{d}x \, \mathrm{d}r \, \mathrm{d}t.
\end{align*}

\textbf{Definition 7} \emph{For $t=0,1$, we define}
\begin{equation*}
m_t(x) := \mathbb{E}[\,Y_t \mid X = x \, ].
\end{equation*}

\textbf{Definition 8} \emph{The variance of $Y_t$ with respect to the distribution $p(x,r,t)$ is given by}
\begin{align*}
 & \sigma_{Y_t}^2(p(x,r,t))  \\
 =\,& \int_{\mathcal{X} \times  \mathcal{R} \times \mathcal{Y}} \!\!\!\!\!\!\!\!\!\!\!\!\!\!\! (y_t - m_t(x))^2 \, p(y_t \mid x) \, p(x,r,t) \, \mathrm{d}y_t \, \mathrm{d}x \,  \mathrm{d}r, 
\end{align*}
\emph{and we define}
\begin{align*}
 & \sigma_{Y_t}^2 = \min\{\sigma_{Y_t}^2(p(x,r,t)), \sigma_{Y_t}^2(p(x,r,1-t))\}, \\
 & \sigma_{Y}^2 = \min\{\sigma_{Y_0}^2, \sigma_{Y_1}^2\}.
\end{align*}

\clearpage

\textbf{Lemma 1} \emph{For any function $f: \mathcal{X} \times \mathcal{T} \rightarrow \mathcal{Y}$ and distribution $p(x,r,t)$ over $\mathcal{X} \times \mathcal{R} \times \mathcal{T}$, we have}
\begin{align*}
&\int_{\mathcal{X} \times \mathcal{R} \times \mathcal{T}} \!\!\!\!\!\!\!\!\!\!\!\!\!\!\! (f_t(x) - m_t(x))^2 \, p(x,r,t) \, \mathrm{d}x \, \mathrm{d}r \, \mathrm{d}t  \\
=\,& \epsilon_{\mathrm{F}}(h, \Phi) - \sigma_{Y_1}^2(p(x,r,T=1)) - \sigma_{Y_0}^2(p(x,r,T=0))  \\
\leq\,& \epsilon_{\mathrm{F}}(h, \Phi) - 2 \, \sigma_{Y}^2
\end{align*}
\emph{and}
\begin{align*}
&  \int_{\mathcal{X} \times \mathcal{R} \times \mathcal{T}}  (f_t(x) - m_t(x))^2 \, p(x,r,1-t) \, \mathrm{d}x \, \mathrm{d}r \, \mathrm{d}t  \\
=\,& \epsilon_{\mathrm{CF}}(h, \Phi) - \sigma_{Y_1}^2(p(x,r,T=0)) - \sigma_{Y_0}^2(p(x,r,T=1))\\
\leq\,&  \epsilon_{\mathrm{CF}}(h, \Phi) - 2 \, \sigma_{Y}^2,
\end{align*}
\emph{where $\epsilon_{\mathrm{F}}(h, \Phi)$ and $\epsilon_{\mathrm{CF}}(h, \Phi)$ are with respect to the squared loss.}

\emph{Proof}.
\begin{align}
& \epsilon_{\mathrm{F}}(h, \Phi)  \nonumber \\
 =& \int_{\mathcal{X} \times \mathcal{R} \times \mathcal{T}} \!\!\!\!\!\!\!\!\!\!\!\!\! l_{h,\Phi}(x,t) \, p(x,r,t) \, \mathrm{d}x \, \mathrm{d}r \, \mathrm{d}t \nonumber  \\
 =& \int_{\mathcal{X} \times \mathcal{R} \times \mathcal{T} \times \mathcal{Y}} \!\!\!\!\!\!\!\!\!\!\!\!\!\!\!\!\!\!\!\! L_Y(y_t, h_t(\Phi(x))) \, p(y_t \mid x) \,  p(x,r,t) \, \mathrm{d}y_t \, \mathrm{d}x \, \mathrm{d}r \, \mathrm{d}t \nonumber  \\
 =& \int_{\mathcal{X} \times \mathcal{R} \times \mathcal{T} \times \mathcal{Y}} \!\!\!\!\!\!\!\!\!\!\!\!\!\!\!\!\!\!\!\! (f_t(x) - y_t)^2 \, p(y_t \mid x) \,  p(x,r,t) \, \mathrm{d}y_t \, \mathrm{d}x \, \mathrm{d}r \, \mathrm{d}t  \label{eq:lemma1_squared_loss}   \\
 =& \int_{\mathcal{X} \times \mathcal{R} \times \mathcal{T} \times \mathcal{Y}} \!\!\!\!\!\!\!\!\!\!\!\!\!\!\!\!\!\!\!\! (f_t(x) - m_t(x))^2 \, p(y_t \mid x) \,  p(x,r,t) \, \mathrm{d}y_t \, \mathrm{d}x \, \mathrm{d}r \, \mathrm{d}t   \label{eq:lemma1_sum} \\ 
& + \int_{\mathcal{X} \times \mathcal{R} \times \mathcal{T} \times \mathcal{Y}}\!\!\!\!\!\!\!\!\!\!\!\!\!\!\!\!\!\!\!\!  (m_t(x) - y_t)^2 \, p(y_t \mid x) \,  p(x,r,t) \, \mathrm{d}y_t \, \mathrm{d}x \, \mathrm{d}r \, \mathrm{d}t  \nonumber \\
& + \int_{\mathcal{X} \times \mathcal{R} \times \mathcal{T} \times \mathcal{Y}} \!\!\!\!\!\!\!\!\!\!\!\!\!\!\!\!\!\!\!\!  2 \, (f_t(x) - m_t(x)) \, (m_t(x) - y_t) + \cdots \nonumber \\
& + p(y_t \mid x) \, p(x,r,t) \, \mathrm{d}y_t  \, \mathrm{d}x \, \mathrm{d}r \, \mathrm{d}t \nonumber \\
 =& \int_{\mathcal{X} \times \mathcal{R} \times \mathcal{T}} \!\!\!\!\!\!\!\!\!\!\!\!\!\! (f_t(x) - m_t(x))^2 \, p(x,r,t) \, \mathrm{d}x \, \mathrm{d}r \, \mathrm{d}t  \label{eq:lemma1_fin} \\
& + \sigma_{Y_1}^2(p(x,r,T=1)) + \sigma_{Y_0}^2(p(x,r,T=0)). \nonumber
\end{align}

We obtain Eq.~(\ref{eq:lemma1_squared_loss}) for the squared loss function $L_Y$ and by using $ h_t(\Phi(x)) = f_t(x)$. Then, Eq.~(\ref{eq:lemma1_fin}) follows from Definition~8 by summing the second term of Eq.~(\ref{eq:lemma1_sum}) over the space $\mathcal{T}$, \ie, for $t=0,1$ and because the third term in Eq.~(\ref{eq:lemma1_sum}) evaluates to zero since $m_t(x) = \int_{\mathcal{Y}} y_t p(y_t \mid x) \, \mathrm{d}y_t$. We show the proof for $\epsilon_{\mathrm{F}}(h, \Phi)$. The proof for $\epsilon_{\mathrm{CF}}(h, \Phi)$ is analogous. 

\clearpage

\textbf{Lemma 2} \emph{(Bound on PEHE loss).}
\begin{equation*}
\epsilon_{\mathrm{PEHE}}(h, \Phi) \leq 2 \, \left( \epsilon_{\mathrm{F}}(h, \Phi) + \epsilon_{\mathrm{CF}}(h, \Phi) - 4\sigma_{Y}^2 \right),
\end{equation*}
\emph{where $\epsilon_{\mathrm{F}}(h, \Phi)$ and $\epsilon_{\mathrm{CF}}(h, \Phi)$ are with respect to the squared loss.}

\emph{Proof}.
\begin{align}
& \epsilon_{\mathrm{PEHE}}(h, \Phi) \nonumber \\
=\,& \int_{\mathcal{X} \times \mathcal{R}} (\hat{\tau}(x) - \tau(x))^2 \, p(x,r) \, \mathrm{d}x \, \mathrm{d}r  \nonumber \\
 =\,& \int_{\mathcal{X} \times \mathcal{R}} \big( (h_1(\Phi(x)) - h_0(\Phi(x))) - \cdots \nonumber \\
& - (m_1(x) - m_0(x)) \big)^2 \, p(x,r) \, \mathrm{d}x \, \mathrm{d}r  \nonumber \\
 =\,& \int_{\mathcal{X} \times \mathcal{R}} \!\!\!\!\!\!\!\!\!\! \big( (f_1(x) - f_0(x)) - (m_1(x) - m_0(x)) \big)^2 \, p(x,r) \, \mathrm{d}x \, \mathrm{d}r  \nonumber \\
 =\,& \int_{\mathcal{X} \times \mathcal{R}} \!\!\!\!\!\!\!\!\!\! \big( (f_1(x) - m_1(x)) + (m_0(x) - f_0(x)) \big)^2 \, p(x,r) \, \mathrm{d}x  \, \mathrm{d}r  \nonumber \\
\leq\,&  2 \!\! \int_{\mathcal{X} \times \mathcal{R}} \!\!\!\!\!\!\!\!\!\! \big((f_1(x) - m_1(x))^2 + (m_0(x) - f_0(x))^2 \big) \, p(x,r) \, \mathrm{d}x  \, \mathrm{d}r \label{eq:lemma2_pythagoras} \\
 =\,& 2 \,\int_{\mathcal{X} \times \mathcal{R}} \!\!\!\!\!\!\!\!\!\! (f_1(x) - m_1(x))^2 \, p(x,r) \, \mathrm{d}x  \, \mathrm{d}r  \nonumber \\
& + 2 \,\int_{\mathcal{X} \times \mathcal{R}} \!\!\!\!\!\!\!\!\!\! (f_0(x) - m_0(x))^2 \, p(x,r) \, \mathrm{d}x  \, \mathrm{d}r  \nonumber \\
 =\,& 2 \int_{\mathcal{X} \times \mathcal{R}} \!\!\!\!\!\!\!\!\!\! (f_1(x) - m_1(x))^2 \, p(x,r,T=1) \, \mathrm{d}x  \, \mathrm{d}r \label{eq:lemma2_p} \\
& + 2 \int_{\mathcal{X} \times \mathcal{R}} \!\!\!\!\!\!\!\!\!\! (f_1(x) - m_1(x))^2 \, p(x,r,T=0) \, \mathrm{d}x  \, \mathrm{d}r  \nonumber \\
& + 2 \int_{\mathcal{X} \times \mathcal{R}} \!\!\!\!\!\!\!\!\!\! (f_0(x) - m_0(x))^2 \, p(x,r,T=1) \, \mathrm{d}x  \, \mathrm{d}r  \nonumber \\
& + 2 \int_{\mathcal{X} \times \mathcal{R}} \!\!\!\!\!\!\!\!\!\! (f_0(x) - m_0(x))^2 \, p(x,r,T=0) \, \mathrm{d}x  \, \mathrm{d}r   \nonumber \\
 =\,& 2 \int_{\mathcal{X} \times \mathcal{R} \times \mathcal{T}} \!\!\!\!\!\!\!\!\!\!\!\!\!\!\! (f_t(x) - m_t(x))^2 \, p(x,r,t) \, \mathrm{d}x  \, \mathrm{d}r  \, \mathrm{d}t  \nonumber \\
& + 2 \int_{\mathcal{X} \times \mathcal{R} \times \mathcal{T}} \!\!\!\!\!\!\!\!\!\!\!\!\!\!\! (f_t(x) - m_t(x))^2 \, p(x,r,1-t) \,  \mathrm{d}x  \, \mathrm{d}r  \, \mathrm{d}t   \nonumber \\
 \leq\,& 2 \, \big(\epsilon_{\mathrm{F}}(h, \Phi) - 2 \, \sigma_{Y}^2\big) + 2 \, \big( \epsilon_{\mathrm{CF}}(h, \Phi) - 2 \, \sigma_{Y}^2 \big) \label{eq:lemma2_lemma1} \\
 =\,& 2 \, \big(\epsilon_{\mathrm{F}}(h, \Phi) + \epsilon_{\mathrm{CF}}(h, \Phi) - 4\sigma_{Y}^2 \big). \nonumber
\end{align}

We have Eq.~(\ref{eq:lemma2_pythagoras}) because $(x+y)^2 \leq 2 \, (x^2 + y^2)$, Eq.~(\ref{eq:lemma2_p}) because $p(x,r) = p(x,r,T=1) + p(x,r,T=0)$, and Eq.~(\ref{eq:lemma2_lemma1}) because of Lemma~1. Note that the bound can be straightforwardly adapted to the case of the absolute loss function by applying the triangle inequality in Eq.~(\ref{eq:lemma2_pythagoras}). In such case, the standard deviation would be replaced by the mean absolute deviation and the multiplying factor would be 1 instead of 2. 

\clearpage

Further, we define the factual and counterfactual loss in the missing and observed treatment domain, as well as the integral probability metric~(IPM). We use superscripts to denote when we condition on a given variable, \eg, $p^{R=0}(x) = p(X = x \mid R=0)$. 

\textbf{Definition 9} \emph{The expected factual and counterfactual losses of $\Phi$ and $h$ in the missing ($R=0$) and observed ($R=1$) treatment domain are given by}
\begin{align*}
\epsilon_{\mathrm{F}}^{R=1}(h, \Phi) &= \int_{\mathcal{X} \times \mathcal{T}} l_{h,\Phi}(x,t) \, p^{R=1}(x,t) \, \mathrm{d}x \, \mathrm{d}t, \\
\epsilon_{\mathrm{F}}^{R=0}(h, \Phi) &= \int_{\mathcal{X} \times \mathcal{T}} l_{h,\Phi}(x,t) \, p^{R=0}(x,t) \, \mathrm{d}x \, \mathrm{d}t, \\
\epsilon_{\mathrm{CF}}^{R=1}(h, \Phi) &= \int_{\mathcal{X} \times \mathcal{T}} l_{h,\Phi}(x,t) \, p^{R=1}(x,1-t) \, \mathrm{d}x \, \mathrm{d}t, \\
\epsilon_{\mathrm{CF}}^{R=0}(h, \Phi) &= \int_{\mathcal{X} \times \mathcal{T}} l_{h,\Phi}(x,t) \, p^{R=0}(x,1-t) \, \mathrm{d}x \, \mathrm{d}t.
\end{align*}

\textbf{Lemma 3}. \emph{Let $v = p(R=0)$. Then, we have}
\begin{align*}
&\epsilon_{\mathrm{F}}(h, \Phi) = (1-v) \, \epsilon_{\mathrm{F}}^{R=1}(h, \Phi) + v \, \epsilon_{\mathrm{F}}^{R=0}(h, \Phi), \\
&\epsilon_{\mathrm{CF}}(h, \Phi) = (1-v) \, \epsilon_{\mathrm{CF}}^{R=1}(h, \Phi) + v \, \epsilon_{\mathrm{CF}}^{R=0}(h, \Phi).
\end{align*}

The proof follows directly from Definition~6 and Definition~9, by noting that $v = p(R=0)$ and $1-v = p(R=1)$. 

\textbf{Definition 10} \emph{Let $G$ be a function family consisting of functions $g: \mathcal{S} \rightarrow \mathbb{R}$. For a pair of distributions $p_1,p_2$ over $\mathcal{S}$, we define the integral probability metric~(IPM) as}
\begin{equation*}
\mathrm{IPM}_G(p_1,p_2) = \sup_{g \in G} \bigg| \int_{\mathcal{S}} g(s) \, (p_1(s) - p_2(s)) \, \mathrm{d}s \, \bigg|.
\end{equation*}
Thus, $\mathrm{IPM}_G(\cdot,\cdot)$ is a pseudo-metric on the space of probability functions over $\mathcal{S}$. For a sufficiently rich function family $G$, $\mathrm{IPM}_G(\cdot,\cdot)$ is a true metric over the corresponding set of probabilities, \ie, $\mathrm{IPM}_G(p_1,p_2) = 0 \Rightarrow p_1 = p_2$.

\clearpage

\textbf{Lemma 4} \emph{Let $\Phi: \mathcal{X} \rightarrow \mathcal{Z}$ be an invertible representation and $\Psi$ its inverse. Let $p_{\Phi}$ be the distribution induced by $\Phi$ over $\mathcal{Z}$. Let $v = p(R=0)$. Let $G$ be a family of functions $k: \mathcal{Z} \rightarrow \mathbb{R}$ and $\mathrm{IPM}_G(\cdot,\cdot)$ the integral probability metric induced by $G$. Let $h_t: \mathcal{Z} \rightarrow \mathcal{Y}$ for $t=0,1$ be a hypothesis. Assume there exists a constant $B_{\Phi} > 0$, such that, for $t=0,1,$ the function $g_{\Phi,h}(z) := \frac{1}{B_{\Phi}} l_{h,\Phi}(\Psi(z),t) \in G$. Then, we have}
\begin{align*}
& \epsilon_{\mathrm{F}}(h, \Phi) + \epsilon_{\mathrm{CF}}(h, \Phi)    \\
\leq \, & \epsilon_{\mathrm{F}}^{R=1}(h, \Phi) + \epsilon_{\mathrm{CF}}^{R=1}(h, \Phi) \,\,  \\
& + 2 \, v \,  B_{\Phi} \, \mathrm{IPM}_G\big(p_{\Phi}^{R=0}(z), p_{\Phi}^{R=1}(z)\big) .
\end{align*}

\emph{Proof}.
\begin{align}
&\epsilon_{\mathrm{F}}(h, \Phi) + \epsilon_{\mathrm{CF}}(h, \Phi) \nonumber \\
=\,& (1-v) \, \epsilon_{\mathrm{F}}^{R=1}(h, \Phi) \label{eq:lemma4_lemma3} \\
& + v \, \epsilon_{\mathrm{F}}^{R=0}(h, \Phi) + (1-v) \, \epsilon_{\mathrm{CF}}^{R=1}(h, \Phi) + v \, \epsilon_{\mathrm{CF}}^{R=0}(h, \Phi)  \nonumber \\
=\,& \epsilon_{\mathrm{F}}^{R=1}(h, \Phi) + \epsilon_{\mathrm{CF}}^{R=1}(h, \Phi)  \nonumber \\
& + v \, (\epsilon_{\mathrm{F}}^{R=0}(h, \Phi) - \epsilon_{\mathrm{F}}^{R=1}(h, \Phi) + \epsilon_{\mathrm{CF}}^{R=0}(h, \Phi) -  \epsilon_{\mathrm{CF}}^{R=1}(h, \Phi)) \nonumber \\
=\,& \epsilon_{\mathrm{F}}^{R=1}(h, \Phi) + \epsilon_{\mathrm{CF}}^{R=1}(h, \Phi) + \label{eq:lemma4_definition9} \\
&  v \, \bigg( \int_{\mathcal{X} \times \mathcal{T}} \!\!\!\!\!\!\!\!\! l_{h,\Phi}(x,t) \, (p^{R=0}(x,t) - p^{R=1}(x,t)) \, \mathrm{d}x \, \mathrm{d}t   \nonumber \\
& \, \, + \int_{\mathcal{X} \times \mathcal{T}} \!\!\!\!\!\!\!\!\! l_{h,\Phi}(x,t) \, (p^{R=0}(x,1-t) - p^{R=1}(x,1-t)) \, \mathrm{d}x \, \mathrm{d}t \bigg)  \nonumber\\
=& \epsilon_{\mathrm{F}}^{R=1}(h, \Phi) + \epsilon_{\mathrm{CF}}^{R=1}(h, \Phi) + \label{eq:lemma4_assumption2} \\
&  v \, \bigg( \int_{\mathcal{X} \times \mathcal{T}} \!\!\!\!\!\!\!\!\! l_{h,\Phi}(x,t) \, p(t \mid x)(p^{R=0}(x) - p^{R=1}(x)) \, \mathrm{d}x \, \mathrm{d}t   \nonumber \\
& \, \,+ \int_{\mathcal{X} \times \mathcal{T}} \!\!\!\!\!\!\!\!\! l_{h,\Phi}(x,t) \, p(1-t \mid x)(p^{R=0}(x) - p^{R=1}(x)) \, \mathrm{d}x \, \mathrm{d}t \bigg)  \nonumber \\
=\,& \epsilon_{\mathrm{F}}^{R=1}(h, \Phi) + \epsilon_{\mathrm{CF}}^{R=1}(h, \Phi) \label{eq:lemma4_sum_1} \\
& + v \,  \int_{\mathcal{X} \times \mathcal{T}} \!\!\!\!\!\!\!\!\! l_{h,\Phi}(x,t) \, (p^{R=0}(x) - p^{R=1}(x)) \, \mathrm{d}x \, \mathrm{d}t   \nonumber \\
=\,& \epsilon_{\mathrm{F}}^{R=1}(h, \Phi) + \epsilon_{\mathrm{CF}}^{R=1}(h, \Phi)  \nonumber \\
& + v \,  \int_{\mathcal{Z} \times \mathcal{T}} \!\!\!\!\!\!\!\!\! l_{h,\Phi}(\Psi(z),t) \, (p_{\Phi}^{R=0}(z) - p_{\Phi}^{R=1}(z)) \, \mathrm{d}z \, \mathrm{d}t   \nonumber \\
=\,& \epsilon_{\mathrm{F}}^{R=1}(h, \Phi) + \epsilon_{\mathrm{CF}}^{R=1}(h, \Phi)  \nonumber \\
& + v \, B_{\Phi} \,  \bigg( \int_{\mathcal{Z}} \frac{1}{B_{\Phi}} l_{h,\Phi}(\Psi(z),1) \, (p_{\Phi}^{R=0}(z) - p_{\Phi}^{R=1}(z)) \, \mathrm{d}z  \nonumber \\
& + \int_{\mathcal{Z}} \frac{1}{B_{\Phi}} l_{h,\Phi}(\Psi(z),0) \, (p_{\Phi}^{R=0}(z) - p_{\Phi}^{R=1}(z)) \, \mathrm{d}z \bigg) \nonumber \\
\leq\,& \epsilon_{\mathrm{F}}^{R=1}(h, \Phi) + \epsilon_{\mathrm{CF}}^{R=1}(h, \Phi)  \nonumber \\
& + 2 \, v \, B_{\Phi} \,  \sup_{g \in G} \bigg| \int_{\mathcal{Z}} g(z) \, (p_{\Phi}^{R=0}(z) - p_{\Phi}^{R=1}(z)) \, \mathrm{d}z \bigg| \nonumber \\
=\,& \epsilon_{\mathrm{F}}^{R=1}(h, \Phi) + \epsilon_{\mathrm{CF}}^{R=1}(h, \Phi)  \nonumber \\
& + 2 \, v \, B_{\Phi} \, \mathrm{IPM}_G(p_{\Phi}^{R=0}(z), p_{\Phi}^{R=1}(z)). \nonumber
\end{align}

Here, Eq.~(\ref{eq:lemma4_lemma3}) follows from Lemma~3, Eq.~(\ref{eq:lemma4_definition9}) uses Definition~9, Eq.~(\ref{eq:lemma4_assumption2}) follows from Assumption~2 since $T$ is independent of $R$ given $X$, and Eq.~(\ref{eq:lemma4_sum_1}) holds true because $p(t \mid x) + p(1-t \mid x) = 1$. The rest of the proof relies on the assumptions in Lemma~4 and Definition~10. 

\clearpage

Next, we define the factual and counterfactual loss in the treated and the control domain, within the observed treatment domain. Subsequently, we provide a bound for the counterfactual loss.

\textbf{Definition 11} \emph{The expected control ($T=0$) and treated ($T=1$) losses in the observed treatment domain are given by}
\begin{align*}
\epsilon_{\mathrm{F}}^{R=1, T=1}(h, \Phi) &= \int_{\mathcal{X} } l_{h,\Phi}(x,1) \, p^{R=1,T=1}(x) \, \mathrm{d}x, \\
\epsilon_{\mathrm{F}}^{R=1, T=0}(h, \Phi) &= \int_{\mathcal{X}} l_{h,\Phi}(x,0) \, p^{R=1,T=0}(x) \, \mathrm{d}x, \\
\epsilon_{\mathrm{CF}}^{R=1, T=1}(h, \Phi) &= \int_{\mathcal{X} } l_{h,\Phi}(x,1) \, p^{R=1,T=0}(x) \, \mathrm{d}x, \\
\epsilon_{\mathrm{CF}}^{R=1, T=0}(h, \Phi) &= \int_{\mathcal{X} } l_{h,\Phi}(x,0) \, p^{R=1,T=1}(x) \, \mathrm{d}x. 
\end{align*}

\textbf{Lemma 5}. \emph{Let $u = p(T=0)$. Then, we have}
\begin{align*}
&\epsilon_{\mathrm{F}}^{R=1}(h, \Phi) = (1-u) \, \epsilon_{\mathrm{F}}^{R=1, T=1}(h, \Phi) + u \,  \epsilon_{\mathrm{F}}^{R=1, T=0}(h, \Phi), \\
&\epsilon_{\mathrm{CF}}^{R=1}(h, \Phi) = u \, \epsilon_{\mathrm{CF}}^{R=1, T=1}(h, \Phi) + (1-u) \, \epsilon_{\mathrm{CF}}^{R=1, T=0}(h, \Phi).
\end{align*}

The proof follows directly from Definition~9 and Definition~11, by noting that $u = p(T=0)$ and $1-u = p(T=1)$.

\clearpage

\textbf{Lemma 6} \emph{Let $L_Y: \mathcal{Y} \times \mathcal{Y} \rightarrow \mathbb{R}_{+}$ be squared loss function. Let $\Phi: \mathcal{X} \rightarrow \mathcal{Z}$ be an invertible representation and $\Psi$ its inverse. Let $p_{\Phi}$ be the distribution induced by $\Phi$ over $\mathcal{Z}$. Let $u = p(T=0)$. Let $G$ be a family of functions $g: \mathcal{Z} \rightarrow \mathbb{R}$ and $\mathrm{IPM}_G(\cdot,\cdot)$ the integral probability metric induced by $G$. Let $h_t: \mathcal{Z} \rightarrow \mathcal{Y}$ for $t=0,1$ be a hypothesis. Assume there exists a constant $B_{\Phi} > 0$, such that, for $t=0,1,$ the function $g_{\Phi,h}(z) := \frac{1}{B_{\Phi}} l_{h,\Phi}(\Psi(z),t) \in G$. Then, we have}
\begin{align*}
& \epsilon_{\mathrm{CF}}^{R=1}(h, \Phi)  \\
\leq \, &  u \,  \epsilon_{\mathrm{F}}^{R=1, T=1}(h, \Phi) + (1-u) \,  \epsilon_{\mathrm{F}}^{R=1, T=0}(h, \Phi) \, \,   \\
& + B_{\Phi} \, \mathrm{IPM}_G\big(p_{\Phi}^{R=1,T=0}(z), p_{\Phi}^{R=1,T=1}(z)\big).
\end{align*}

\emph{Proof}.
\begin{align}
&\epsilon_{\mathrm{CF}}^{R=1}(h, \Phi) - \big( u \epsilon_{\mathrm{F}}^{R=1, T=1}(h, \Phi) + (1-u) \epsilon_{\mathrm{F}}^{R=1, T=0}(h, \Phi)\big)  \nonumber  \\
=\,& \big( u  \epsilon_{\mathrm{CF}}^{R=1, T=1}(h, \Phi) + (1-u)  \epsilon_{\mathrm{CF}}^{R=1, T=0}(h, \Phi)\big) \label{eq:lemma6_lemma5} \\
& - \big( u \epsilon_{\mathrm{F}}^{R=1, T=1}(h, \Phi) + (1-u) \epsilon_{\mathrm{F}}^{R=1, T=0}(h, \Phi)\big)  \nonumber \\
=\,& u \big(\epsilon_{\mathrm{CF}}^{R=1, T=1}(h, \Phi) - \epsilon_{\mathrm{F}}^{R=1, T=1}(h, \Phi) \big)  \nonumber \\
& + (1-u) \big(\epsilon_{\mathrm{CF}}^{R=1, T=0}(h, \Phi) - \epsilon_{\mathrm{F}}^{R=1, T=0}(h, \Phi) \big)  \nonumber  \\
=\,& u \int_{\mathcal{X} } l_{h,\Phi}(x,1) \, (p^{R=1,T=0}(x) - p^{R=1,T=1}(x)) \, \mathrm{d}x \label{eq:lemma6_definition11} \\
& + (1-u) \int_{\mathcal{X} } l_{h,\Phi}(x,0) \, (p^{R=1,T=1}(x) - p^{R=1,T=0}(x)) \, \mathrm{d}x  \nonumber \\
=\,&  u \int_{\mathcal{Z} } l_{h,\Phi}(\Psi(z),1) \, (p_{\Phi}^{R=1,T=0}(z) - p_{\Phi}^{R=1,T=1}(z)) \, \mathrm{d}z  \nonumber \\
& + (1-u) \int_{\mathcal{Z} } l_{h,\Phi}(\Psi(z),0) \, (p_{\Phi}^{R=1,T=1}(z) - p_{\Phi}^{R=1,T=0}(z)) \, \mathrm{d}z  \nonumber 
\end{align}
\begin{align}
=\,& B_{\Phi} \, u \, \int_{\mathcal{Z} } \frac{1}{B_{\Phi}} l_{h,\Phi}(\Psi(z),1) \, (p_{\Phi}^{R=1,T=0}(z) - p_{\Phi}^{R=1,T=1}(z)) \, \mathrm{d}z  \nonumber \\
& + B_{\Phi} \, (1-u) \,\int_{\mathcal{Z} } \frac{1}{B_{\Phi}} l_{h,\Phi}(\Psi(z),0) \cdots \nonumber \\
& \qquad\qquad \qquad \qquad   (p_{\Phi}^{R=1,T=1}(z) - p_{\Phi}^{R=1,T=0}(z)) \, \mathrm{d}z  \nonumber  \\
\leq\,& B_{\Phi}\, u \, \sup_{g \in G} \bigg| \int_{\mathcal{Z}} g(z) \, (p_{\Phi}^{R=1,T=0}(z) - p_{\Phi}^{R=1,T=1}(z)) \, \mathrm{d}z \bigg|  \nonumber \\
& + B_{\Phi} \, (1-u) \, \sup_{g \in G} \bigg| \int_{\mathcal{Z}} g(z) \, (p_{\Phi}^{R=1,T=1}(z) - p_{\Phi}^{R=1,T=0}(z)) \, \mathrm{d}z \bigg|  \nonumber \\
=\,& B_{\Phi} \, \mathrm{IPM}_G(p_{\Phi}^{R=1,T=0}(z), p_{\Phi}^{R=1,T=1}(z)). \nonumber
\end{align}

Eq.~(\ref{eq:lemma6_lemma5}) follows from Lemma~5, Eq.~(\ref{eq:lemma6_definition11}) is given by Definition~11, and the rest of the proof relies on assumptions in Lemma~6 and Definition~10. Next, we provide the main result of this paper in Theorem~1. 

\clearpage

\textbf{Theorem 1} \emph{Let $\Phi: \mathcal{X} \rightarrow \mathcal{Z}$ be an invertible representation and $\Psi$ its inverse. Let $p_{\Phi}$ be the distribution induced by $\Phi$ over $\mathcal{Z}$. Let $v = p(R=0)$. Let $G$ be a family of functions $g: \mathcal{Z} \rightarrow \mathbb{R}$ and $\mathrm{IPM}_G(\cdot,\cdot)$ the integral probability metric induced by $G$. Let $h_t: \mathcal{Z} \rightarrow \mathcal{Y}$ for $t=0,1$ be a hypothesis. Assume there exists a constant $B_{\Phi} > 0$, such that, for $t=0,1,$ the function $g_{\Phi,h}(z) := \frac{1}{B_{\Phi}} l_{h,\Phi}(\Psi(z),t) \in G$. Then, we have}
\begin{align}
& \epsilon_{\mathrm{PEHE}}(h, \Phi) \nonumber \\
\leq \, &  2 \, (\epsilon_{\mathrm{F}}(h, \Phi) + \epsilon_{\mathrm{CF}}(h, \Phi) - 4\sigma_{Y}^2) \label{eq:theorem1_lemma2} \\
\leq \, &  2 \, \big(\epsilon_{\mathrm{F}}^{R=1}(h, \Phi) + \epsilon_{\mathrm{CF}}^{R=1}(h, \Phi) \label{eq:theorem1_lemma4} \\
& + 2 \, v \, B_{\Phi} \, \mathrm{IPM}_G(p_{\Phi}^{R=0}(z), p_{\Phi}^{R=1}(z)) - 4\sigma_{Y}^2 \big) \nonumber \\
\leq \, & 2 \, \bigg[\epsilon_{\mathrm{F}}^{R=1, T=1}(h, \Phi) + \epsilon_{\mathrm{F}}^{R=1, T=0}(h, \Phi) \label{eq:theorem1_lemma6} \\
&  \quad + B_{\Phi} \, \mathrm{IPM}_G(p_{\Phi}^{R=1, T=0}(z), p_{\Phi}^{R=1,T=1}(z))  \nonumber \\
&  \quad + 2 \, v \, B_{\Phi} \, \mathrm{IPM}_G(p_{\Phi}^{R=0}(z), p_{\Phi}^{R=1}(z)) - 4\sigma_{Y}^2\bigg]. \nonumber 
\end{align}
\emph{where $\epsilon_{\mathrm{F}}(h, \Phi)$ and $\epsilon_{\mathrm{CF}}(h, \Phi)$ are with respect to the squared loss.}

The proof follows directly from Lemma~2, Lemma~4, and Lemma~6. Eq.~(\ref{eq:theorem1_lemma2}) follows from Lemma~2, Eq.~(\ref{eq:theorem1_lemma4}) follows from Lemma~4, and Eq.~(\ref{eq:theorem1_lemma6}) follows from Lemma~6 and by observing that
\begin{align*}
& \epsilon_{\mathrm{F}}^{R=1}(h, \Phi) + u \, \epsilon_{\mathrm{F}}^{R=1, T=1}(h, \Phi) + (1-u) \, \epsilon_{\mathrm{F}}^{R=1, T=0}(h, \Phi) \\
=\,& (1-u) \, \epsilon_{\mathrm{F}}^{R=1, T=1}(h, \Phi) + u \, \epsilon_{\mathrm{F}}^{R=1, T=0}(h, \Phi)  \\
& + u \, \epsilon_{\mathrm{F}}^{R=1, T=1}(h, \Phi) + (1-u) \, \epsilon_{\mathrm{F}}^{R=1, T=0}(h, \Phi)  \\
=\,& \epsilon_{\mathrm{F}}^{R=1, T=1}(h, \Phi) + \epsilon_{\mathrm{F}}^{R=1, T=0}(h, \Phi).
\end{align*}

\newpage

\section{IMPLEMENTATION DETAILS}
\label{app:exp_det}

\textbf{Hyperparameters.} The hyperparameters for our MTRNet and the two other deep learning baselines, \ie, TARNet and CFRMMD, include: representation layer size, hypothesis layer size, number of iterations, batch size, learning rate, dropout rate, $\lambda$, $\alpha$ (only for MTRNet and CFRMMD), and $\beta$ (only for MTRNet). We used large similar tuning ranges for datasets (an exception is the batch size, which we varied to reflect the different sizes of the datasets). 
\begin{itemize}
\item \textbf{IHDP}. Here, we have: representation layer size $\in \{50, 100, 200\}$, hypothesis layer size $\in \{50, 100, 200\}$, number of iterations $\in \{100, 200, 300\}$, batch size $\in \{50, 70, 100\}$, learning rate $\in \{0.01, 0.005, 0.001, 0.0005, 0.0001\}$, dropout rate $\in  \{0.1, 0.2, 0.3\}$, $\lambda \in \{0.0005, 0.0001, 0.00005\}$, $\alpha \in \{10^{k/2}\}_{k=-4}^{2}$, and $\beta \in \{10^{k/2}\}_{k=-4}^{2}$. 
\item \textbf{Twins}. Here, we have: batch size $\in \{500, 1000, 1500\}$. The rest of the hyperparameter ranges are the same as for IHDP.
\item \textbf{Jobs}. Here, we have: batch size $\in \{200, 300, 500\}$. The rest of the hyperparameter ranges are the same as for IHDP.
\end{itemize}

\textbf{Cross-validation.} For real-world data, the standard cross-validation cannot be used with the PEHE loss because we observe only the factual outcome, which means that we do not have access to CATE. However, we can compute a substitute for CATE by using the nearest neighbor in the opposite treatment group as a surrogate for the counterfactual outcome. Hence, to compute a substitute for CATE for a data point $i$, we use the factual outcome $y_i$ and a surrogate for the counterfactual outcome $y_{j(i)}$, where $j(i)$ is the nearest neighbor of $i$ in the opposite treatment group, \ie, $t_{j(i)} = 1 - t_i$. Then, we have the nearest neighbor approximation of the PEHE loss given by $\hat{\epsilon}_{\mathrm{{PEHE}_\mathrm{nn}}} = \frac{1}{n} \sum_{i=1}^n \big(\hat{\tau}(x) - (1-2t_i) \, (y_{j(i)} - y_1) \big)^2$. We use $\hat{\epsilon}_{\mathrm{{PEHE}_\mathrm{nn}}}$ for hyperparameter selection via cross-validation for IHDP and Twins. For Jobs, we directly use the policy risk for cross-validation. 

\textbf{Data pre-processing.} In our experiments, we modify the datasets such that treatment information is partially missing. The missingness mechanism is designed such that treatment missingness $R$ depends on covariates $X$. We do this in the following way. For each data point $i$, we have the probability of missingness $p_{m(i)}$, and the probability that the treatment is observed $p_{o(i)}$. Initially, we set them both to 1. Then, for a data point $i$ and covariate $X_j$, if $x_{ji}$ is larger than the empirical mean of $X_j$, we multiply $p_{m(i)}$ with parameter $q \in (0,1)$, and $p_{o(i)}$ with $1-q$. If $x_{ji}$ is smaller than the empirical mean of $X_j$, we multiply $p_{m(i)}$ with $1-q$ and $p_{o(i)}$ with $q$. We iterate the procedure over all data points $i = 1, \ldots ,n$, for each covariate $j = 1, \ldots, d$. Following this, we normalize $p_{m(i)}$ and $p_{o(i)}$ by dividing each with their sum which gives us the probability of treatment missingness for every data point $i$. Then, we randomly sample $r_i$ from the set $\{0,1\}$ such that zero is sampled with probability $p_{m(i)}$, and one is sampled with probability $p_{o(i)} = 1 - p_{m(i)}$. In the end, we control the overall proportion of missing treatments using parameter $m \in (0,1)$ by randomly changing some $r_i$ such that the final proportion of missing treatments equals $m$. Hence, $m$ controls the overall probability of treatment missingness, and $q$ controls the magnitude of the covariate shift between the observed and missing treatment population. Here: the further away $q$ is from $0.5$, the larger is the covariate shift. 

\newpage
\section{ADDITIONAL EXPERIMENTS}
\label{app:add_exp}

\begin{figure}[H]
\centerline{\includegraphics[width= 9cm]{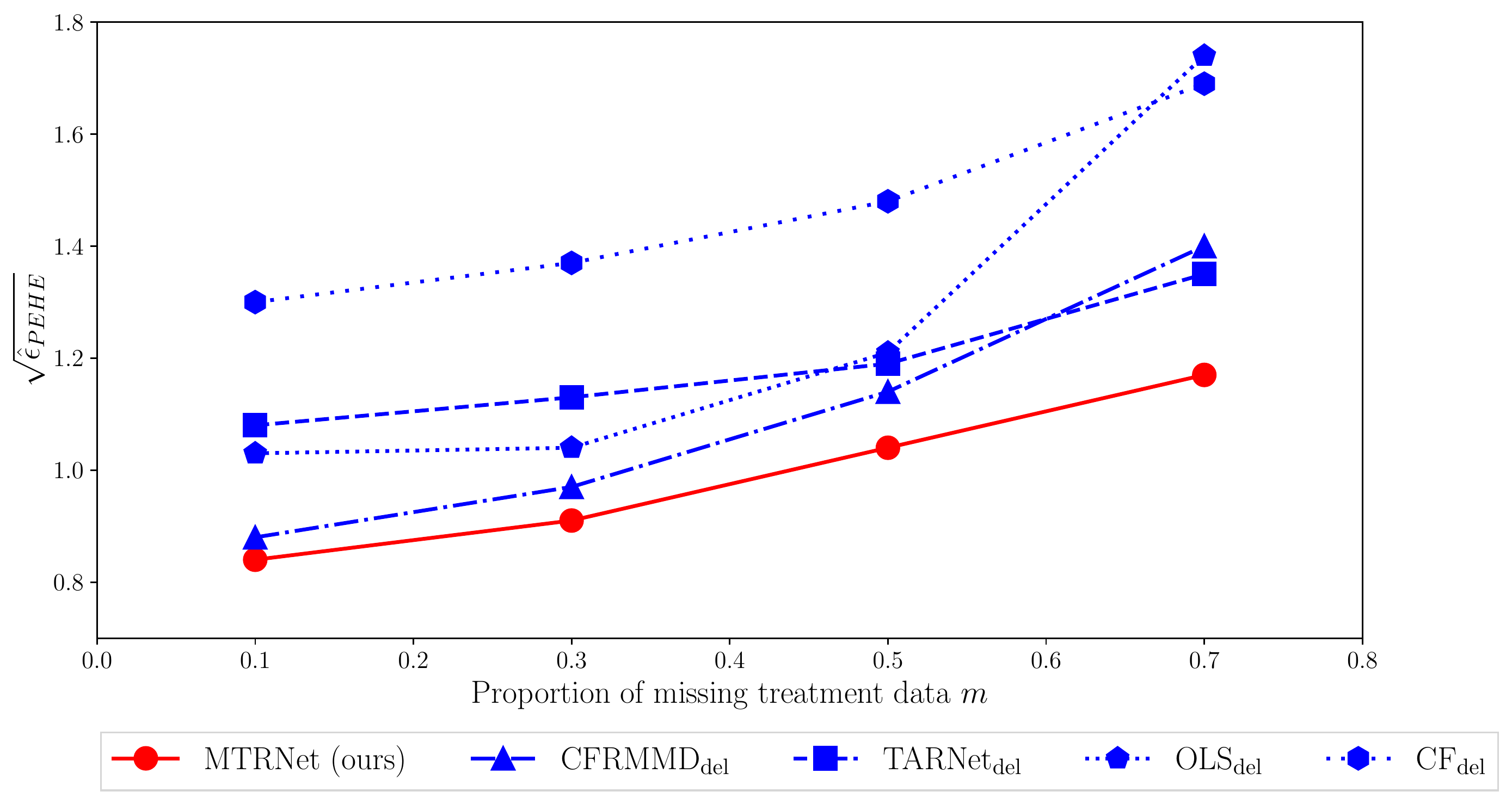}}
\caption{IHDP results for increasing proportion, $m$, of missing treatment data (here: deletion method ``del'' for handling missing treatments).}
\label{fig:m_results2}
\end{figure}

\begin{figure}[H]
\centerline{\includegraphics[width= 9cm]{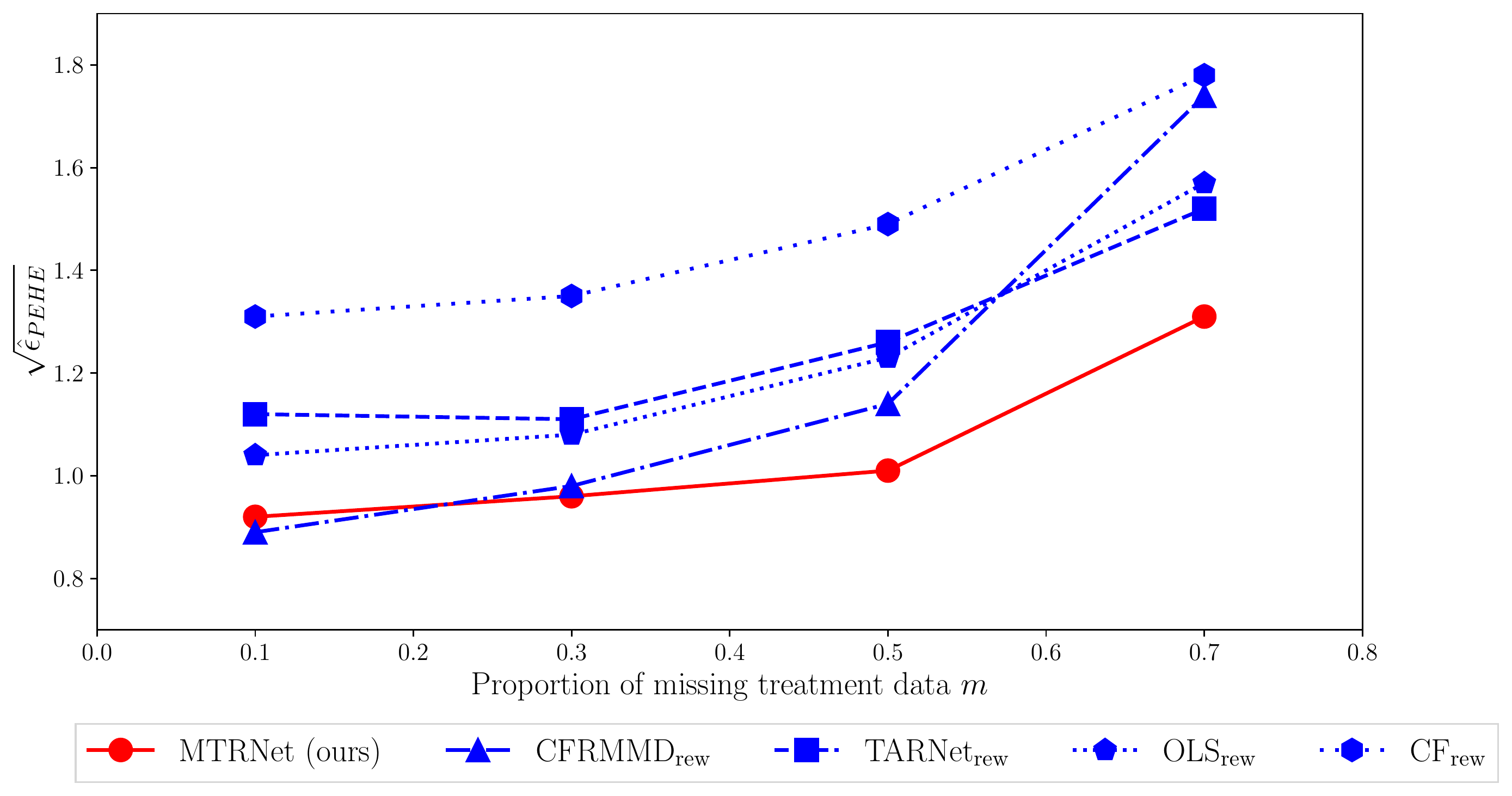}}
\caption{IHDP results for increasing proportion, $m$, of missing treatment data (here: re-weighting method ``rew'' for handling missing treatments).}
\label{fig:m_results3}
\end{figure}

Here, we show the results of IHDP experiments when varying the parameter $m$, \ie, the proportion of missing treatment data. The results are shown for MTRNet and the four CATE estimation methods with different methods for handling missing treatments, namely, with deletion method in Fig.~\ref{fig:m_results2}, and with re-weighting method in Fig.~\ref{fig:m_results3}. We confirm the finding from our main paper that the performance gap between our MTRNet and the baseline methods becomes larger as we increase the proportion of missing treatment data, \ie, the parameter $m$. 

\end{document}